\documentclass[runningheads]{llncs}
\usepackage{graphicx}
\usepackage{cvpr}
\usepackage{amsmath,amssymb} %
\usepackage{color}

\usepackage{epsfig}
\usepackage{graphicx}
\usepackage{amsmath}
\usepackage{amssymb}

\usepackage{graphics}
\usepackage{pifont}
\usepackage{color}
\usepackage{booktabs}
\usepackage{multirow}
\usepackage{subcaption}
\usepackage{gensymb}
\usepackage{bm}
\usepackage{algorithm} 
\usepackage{algpseudocode} 
\usepackage{capt-of}
\usepackage[table]{xcolor}
\usepackage{footnote}

\makesavenoteenv{tabular}
\makesavenoteenv{table}

\usepackage[pagebackref=true,breaklinks=true,letterpaper=true,colorlinks,bookmarks=false]{hyperref}
\begin{document}

\pagestyle{headings}
\mainmatter
\def\ECCVSubNumber{3099}  %

\title{Deep Feedback Inverse Problem Solver} %

\title{Deep Feedback Inverse Problem Solver}
\titlerunning{Deep Optimizer} 
\authorrunning{\textcolor{black}{\href{http://people.csail.mit.edu/weichium/}{\textcolor{black}{Wei-Chiu Ma}}} \etal} 
\author{\textcolor{black}{\href{http://people.csail.mit.edu/weichium/}{\textcolor{black}{Wei-Chiu Ma$^{1,2}$}}} \quad Shenlong Wang$^{1,3}$ \quad Jiayuan Gu$^{1,4}$ \quad Siva Manivasagam$^{1,3}$ \quad Antonio Torralba$^{2}$ \quad Raquel Urtasun$^{1,3}$}
\institute{
$^{1}$Uber Advanced Technologies Group \quad $^{2}$Massachusetts Institute of Technology\\
$^{3}$University of Toronto  \quad $^{4}$University of California San Diego%
}

\newcommand{\shenlong}[1]{\textcolor{black}{ #1}}
\newcommand{\slwang}[1]{\textcolor{blue}{Shenlong: #1}}
\newcommand{\raquel}[1]{\textcolor{red}{Raquel: #1}}
\newcommand{\weichiu}[1]{\textcolor{blue}{Wei-Chiu: #1}}
\newcommand{\wcma}[1]{\textcolor{black}{#1}}
\newcommand{\todo}[1]{\textcolor{blue}{}}
\newcommand{\cbd}[1]{\textcolor{black}{#1}}
\newcommand{\addon}[1]{\textcolor{black}{#1}}

\newcommand{\arxiv}[1]{\vspace{0mm}}
\newcommand{\arxivscale}[1]{\scalebox{#1}}
\newcommand{\bx}{\mathbf{x}}
\newcommand{\bb}{\mathbf{b}}
\newcommand{\ba}{\mathbf{a}}
\newcommand{\bo}{\mathbf{o}}
\newcommand{\bz}{\mathbf{z}}
\newcommand{\bp}{{\bm{p}}}
\newcommand{\bq}{{\bm{q}}}
\newcommand{\bn}{\mathbf{n}}
\newcommand{\bw}{\mathbf{w}}
\newcommand{\cI}{\mathcal{I}}
\newcommand{\cF}{\mathcal{F}}
\newcommand{\bK}{\mathbf{K}}
\newcommand{\bR}{\mathbf{R}}
\newcommand{\bW}{\mathbf{W}}
\newcommand{\bJ}{\mathbf{J}}
\newcommand{\cL}{\mathcal{L}}
\newcommand{\cS}{\mathcal{S}}
\newcommand{\cD}{\mathcal{D}}
\newcommand{\cR}{\mathcal{R}}
\newcommand{\cG}{\mathcal{G}}
\newcommand{\cB}{\mathcal{B}}
\newcommand{\bbR}{\mathbb{R}}
\newcommand{\cM}{\mathcal{M}}
\newcommand{\by}{\mathbf{y}}
\newcommand{\ut}{^{(t)}}
\newcommand{\up}{^{(t-1)}}
\newcommand{\bt}{\mathbf{t}}
\newcommand{\bxi}{\bm{\xi}}
\newcommand{\btheta}{\bm{\theta}}
\newcommand{\bpsi}{\bm{\psi}}
\newcommand{\bcM}{\bm{\mathcal{M}}}

\maketitle

\begin{abstract}
We present an efficient, effective, and generic approach towards solving inverse problems. The key idea is to leverage the feedback signal provided by the forward process and learn an iterative update model.
Specifically, at each iteration, the neural network takes the feedback as input and outputs an update on current estimation. Our approach does not have any restrictions on the forward process; it does not require any prior knowledge either. Through the feedback information, our model not only can produce accurate estimations that are coherent to the input observation but also is capable of recovering from early incorrect predictions. We verify the performance of our approach over a wide range of inverse problems, including 6-DOF pose estimation, illumination estimation, as well as inverse kinematics. Comparing to traditional optimization-based methods, we can achieve comparable or better performance while being two to three orders of magnitude faster. Compared to deep learning-based approaches, our model consistently improves the performance on all metrics. Please refer to the \href{http://people.csail.mit.edu/weichium/papers/eccv20-deep-optimizer/}{project page} for videos, animations, supplementary materials, etc.
\end{abstract}

%
\section{Introduction}
Given a 3D model of an object, the light source(s), and their relevant poses to the camera, one can generate highly realistic images of the scene with one click. %
While such a \emph{forward} rendering process is complicated and requires explicit modeling of interreflections, self-occlusions, as well as distortions, it is well-defined and can be computed effectively. 
However, if we were to recover the illumination parameters or predict the 6 DoF pose of the object from the image in an \emph{inverse} fashion, the task becomes extremely challenging. This is because a lot of crucial information is lost during the forward (rendering) process. %
In fact, many complicated tasks in natural science, signal processing, and robotics, all face similar challenges -- the model parameters of interest cannot be measured directly and need to be estimated from limited observations. 
{This family of problems are commonly referred to as \textbf{inverse problems}.}
Unfortunately, while there exists sophisticated theories on how to design the forward processes, %
how to address the inherent ambiguities of the inverse problem still remains an open question. 

{One popular strategy to disambiguate the solution is to model the inverse problem as a structured optimization task and incorporate human knowledge into the model \cite{he2010single,pan2016blind,huang2015single,rother2011recovering}. For instance, the estimated solution should agree with the observations \cite{pan2016l_0} and be smooth \cite{oh2001image,bell2014intrinsic}, or should follow a certain statistical distribution \cite{levin2007blind,wang2015learning,barron2014shape}.}
Through imposing carefully designed objectives, classic structure optimization methods are able to 
{find a solution that not only agrees with the observations but also satisfies our prior knowledge about the solution.}
In practice, however, almost no hand-crafted priors can succeed in including all phenomena. 
To ensure that the optimization problem can be solved efficiently, there are multiple restrictions on the form of these priors as well as the the forward process \cite{boyd2004convex}, both of which increase the difficulty of design. 
{Furthermore, most optimization approaches require many iterations to converge and are sensitive to initialization.} 

\begin{figure}[tb]
\centering
\includegraphics[width=0.9\linewidth,trim={20mm 123mm 121mm 7mm},clip]{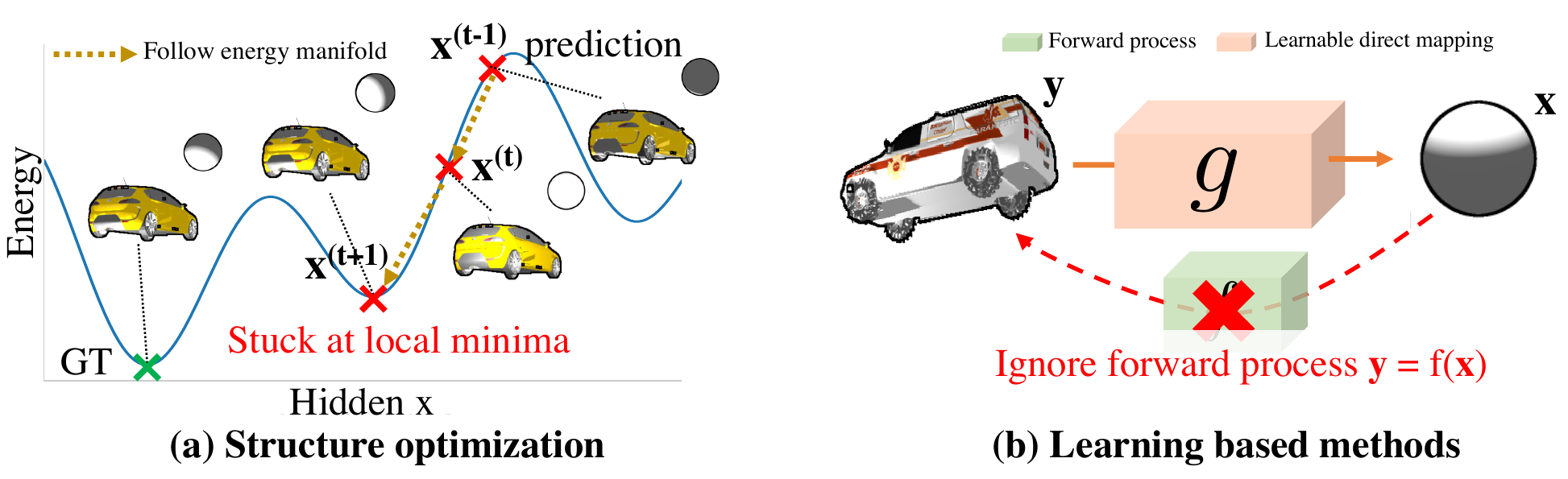}
\arxiv{-3mm}
\caption{\textbf{Prior work on inverse problems}: \wcma{(a) Structured optimization approaches require hand-crafted energy/objective functions and are sensitive to initializations which makes them easy to get stuck in local optima. (b) Direct learning based methods do not utilize the available forward process as feedback to guarantee the quality of the solution. Without this feedback, the models cannot rectify the estimates effectively as shown above.}
}
\arxiv{-6mm}
\end{figure}

On the other hand, learning based methods propose to directly learn a mapping from observations to the model parameters \cite{tung2017self,kanazawa2018learning,yao20183d,wu2016physics,rick2017one}. {They capitalize on powerful machine learning tools to extract task-specific priors in a data-driven fashion.}
With the help of large-scale datasets and the flourishing of deep learning, they are able to achieve state-of-the-art performance on a variety of inverse problems \cite{wu2015galileo,tung2017adversarial,shocher2018zero,epstein2019oops,kanazawa2018end,ma2018single}.  
Unfortunately, these methods often ignore the fact that the forward model for inverse problems is available. Their systems remain open loop and do not have the capability to \emph{update} their prediction based on the \emph{feedback signal}. {Consequently, the estimated parameters, while performing well in the majority of the cases, may generate results that are either incompatible with the real observations or not realistic.}
With these challenges in mind, we develop a novel approach to solving inverse problems that takes the best of both worlds. 
{The key idea is to \emph{learn to iteratively update} the current estimation through the \emph{feedback signal} from the forward process. Specifically, we design a neural network that takes the observation and the forward simulation result of the previous estimation as input, and outputs a steep update towards the ideal model parameters.}{
The advantages are four-fold:
First, as each update is trained to aggressively move towards the ground truth, we can accelerate the update procedure and reach the target with much fewer iterations than classic optimization approaches. 
Second, our approach does not need to explicitly define the energy. 
Third, we do not have any restrictions on the forward process, such as differentiability, which greatly expands the applicable domain. 
Finally, in contrast to  conventional learning methods, %
our method incorporates feedback signals from the forward process so that the network is aware of how close the current estimation is to the ground truth and can react accordingly. The estimated parameters generally lead to results closer to the observation.}

We demonstrate the effectiveness of our approach on three different inverse problems in graphics and robotics: illumination estimation, 6 DoF pose estimation, and inverse kinematics. Compared to traditional optimization based methods, we are able to achieve comparable or better performance while being two to three orders of magnitude faster. Compared to deep learning based approaches, our model consistently improves the performance on all metrics.

%

\begin{figure}[tb]
\centering
\includegraphics[width=0.95\linewidth,trim={5mm 40mm 100mm 3mm},clip]{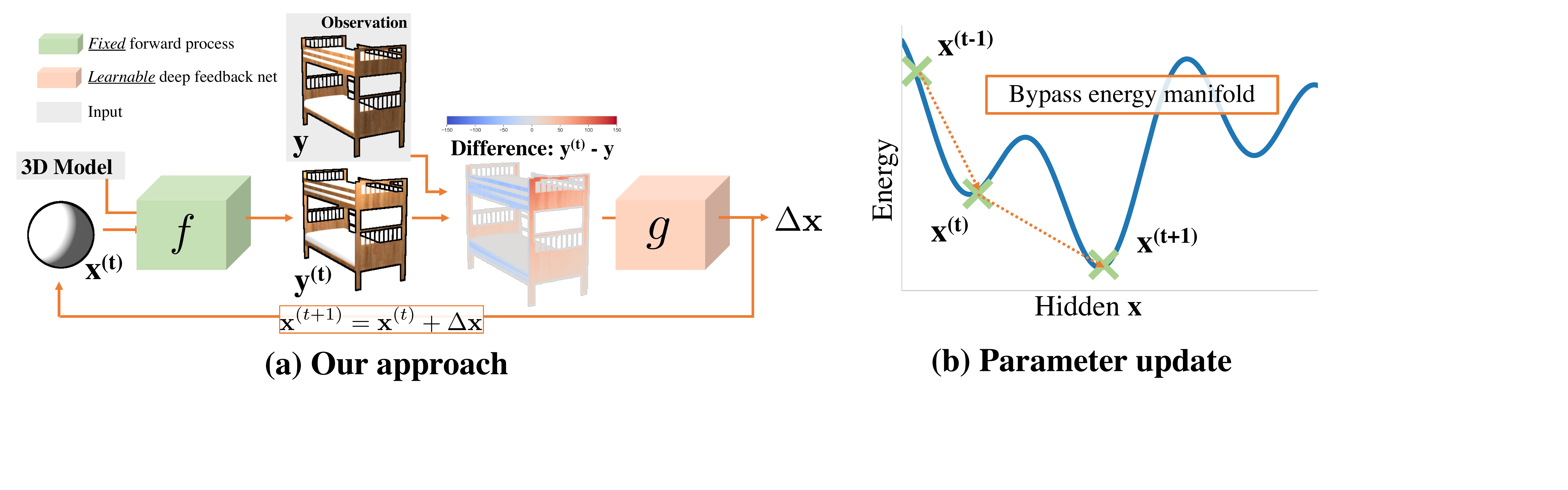}
\arxiv{-3mm}
\caption{
\textbf{Overview:} Our model {iteratively updates} the estimation based on the {feedback signal} from the forward process. We adopt a closed-loop scheme to ensure the consistency between the estimation and the observation. We neither require an objective at test time, nor have any restrictions on the forward process. Click \href{http://people.csail.mit.edu/weichium/img/deep-optimizer-teaser.gif}{here} to watch an animated version of the update procedure.
}
\arxiv{-3mm}
\end{figure}

\arxiv{-3mm}
\section{Background}
\arxiv{-3mm}
\label{sec:preliminary}
Let $\bx \in \mathcal{X}$ be the hidden parameters of interest and let $\by \in \mathcal{Y}$ be the measurable observations. Denote $f: \bx\rightarrow\by$ as the deterministic forward process. 
{The aim of inverse problem is to recover $\bx$ given the observation $\by$ and the forward mapping $f$.} In the tasks that we consider, $\mathcal{X}$ is a group such as $\mathcal{X} = \text{SE(3)}$ for 6 DoF pose estimation and $\mathcal{X} = \mathbb{R}^3$ when estimating the position of the light source

\subsection{Structured optimization}
\wcma{Structured optimization methods generally formulate the inverse problem as an energy minimization task \cite{chan2006total,dong2011image,donoho1995noising,portilla2003image,laffont2015intrinsic,pan2016l_0,huang2015single}: %
\begin{equation*}
\bx^\ast = \arg\min_\bx E(\bx) = \arg\min_\bx E_\mathrm{data}(f(\bx), \by) + \lambda E_\mathrm{prior}(\bx),
\end{equation*}
where the data term $E_\mathrm{data}$ measures the similarity between the observation $\by$ and the forward simulated results $f(\bx)$ of the hidden parameters $\bx$; and the prior term $E_\mathrm{prior}$ encodes humans' knowledge about {the solution $\bx$}.
} 
As the energy function is often non-convex, iterative algorithms are used to refine the estimation. {Without loss of generality, the update rule can be written as: }
{
\begin{equation}
\label{equ:grad}
\bx^{t+1} = \bx^t + g_E(\bx^t, \by^t, \by)
\end{equation}
where $g_E(\bx^t, \by^t, \by)$ is an analytical update function derived from the energy function $E$, and $\by^t = f(\bx^t)$. For instance, in continuous-valued inverse problems, $g_E = - A_{E}(\bx^t) \nabla_E (\bx^t) $}, where $\nabla_E(\bx)$ is the first-order Jacobian and $A_E$ is a warping matrix that depends on the optimization algorithm and the form of the energy.
For instance, $A_E$ is simply a (approximated) Hessian matrix in Newton method and is equivalent to the step size in first order gradient descent. 

One major advantage of these approaches is that they explicitly take into account how close $f(\bx)$ and $\by$ are via the data term $E_\mathrm{data}$, and exploit such \emph{feedback} as a guidance for the update. 
{This ensures that the result $f(\bx^\ast)$ generated from the final estimation $\bx^\ast$ is close to the observation $\by$.}
While impressive results have been achieved, there are several challenges remaining: first, they require {both the forward process $f$ as well as the prior $E_\mathrm{prior}$} to be {optimization-friendly} (\eg differentiable) so that {inference algorithms} can be applied. {Unfortunately this is not the case for many inverse problems and tailored approximations are required \cite{kato2018neural,liu2019soft,ravi2020pytorch3d,lin2019photometric,wiles2019synsin}. The performance may thus be affected.}
Second, they often require many updates to reach {a decent solution} (\eg first-order methods). 
If higher order methods are exploited to speed up the process, the update may become expensive (\eg, second-order methods). %
Third, carefully designed priors are necessary for identifying the true solution from multiple feasible answers. {This is particularly true for ill-posed inverse problems, such as super-resolution and inverse kinematics, in which there exists infinite number of feasible solutions that could generate the observation.} 
Additionally, the energy must be designed in a way that is easy to optimize, which is sometimes non-trivial. 
Finally, these optimization methods are typically sensitive to the initialization. 

\begin{algorithm}[t]
\caption{Deep Feedback Inverse Problem Solver}
\label{alg:dso}
\begin{algorithmic}[1]
\State \textbf{input} observation $\by$, forward model $f(\cdot)$ and init $\bx^0$
\For {$iter=0,1,\ldots, T-1$}
\State Run forward model: $\by^t = f(\bx^t)$
\State Compute update: $\bx^{t+1} = \bx^{t} + g_\bw(\bx^t, \by^t, \by)$
\EndFor
\State \textbf{output} $\bx^T$
\end{algorithmic}
\end{algorithm}

\arxiv{-3mm}
\subsection{Learning based methods}
Another line of work \cite{dong2014learning,xu2014deep,ledig2017photo,lin2018learning,lai2017deep} has been devoted to directly learning a mapping from the observations $\by$ to the solution $\bx$:
\begin{equation}
\label{equ:learning}
\bx^\ast = g(\by; \bw).
\end{equation}
Here, $g(\cdot; \bw)$ is a learnable function parameterized by $\bw$. 
These approaches try to capitalize on the feature learning capabilities of deep neural networks to extract statistical priors from data, and approximate the inverse process without the help of any hand-crafted energies. While these methods have achieved state-of-the-art performance in many challenging inverse tasks such as inverse kinematics \cite{pavllo:quaternet:2018,zhou2019continuity}, super-resolution \cite{ledig2017photo,wang2018esrgan}, compressive sensing \cite{kulkarni2016reconnet}, image inpainting \cite{pathak2016context,liu2018image}, illumination estimation \cite{BigTimeLi18,ma2018single}, reflection separation \cite{zhang2018single}, and image deblurring \cite{nah2017deep}, they ignore the fact that the forward process $f$ is known. 

{As a consequence, there} is no \emph{feedback} mechanism within the model that scores if $f(\bx^\ast)$ is close to $\by$ after the inference, and the model cannot update the estimation accordingly. The whole system remains \emph{open loop}.
\todo{re-write last sentence}
%
%
%
%
%

%

%

%
%
%
%
%
%
%
%
%

%
%
%
%
%
%
%
%
%
%
%
%
%

%
%
%
%
%

%
%
%

%


%
\section{Deep Feedback Inverse Problem Solver}
In this paper we aim  to develop an extremely efficient yet effective approach to solving structured inverse problems. We build our model based on the observation that traditional optimization approaches and current learning based methods are complementary -- one is good at {exploiting feedback signals as guidance and inducing human priors} %
, while the other excels at {learning data-driven inverse mapping.}
Towards this goal, we present a simple solution that takes the best of both worlds. %
{We first describe our deep feedback network that iteratively updates the solution based on the feedback signal generated by the forward process. Then we demonstrate how to perform efficient inference as well as learning. Finally, we discuss our design choices and the relationships to related work.}

\begin{figure*}[tb]
\centering
\setlength{\tabcolsep}{1pt}
\arxivscale{1}{
\begin{tabular}{ccc}
\includegraphics[width=0.33\linewidth,trim={3mm 4mm 3mm 0mm},clip]{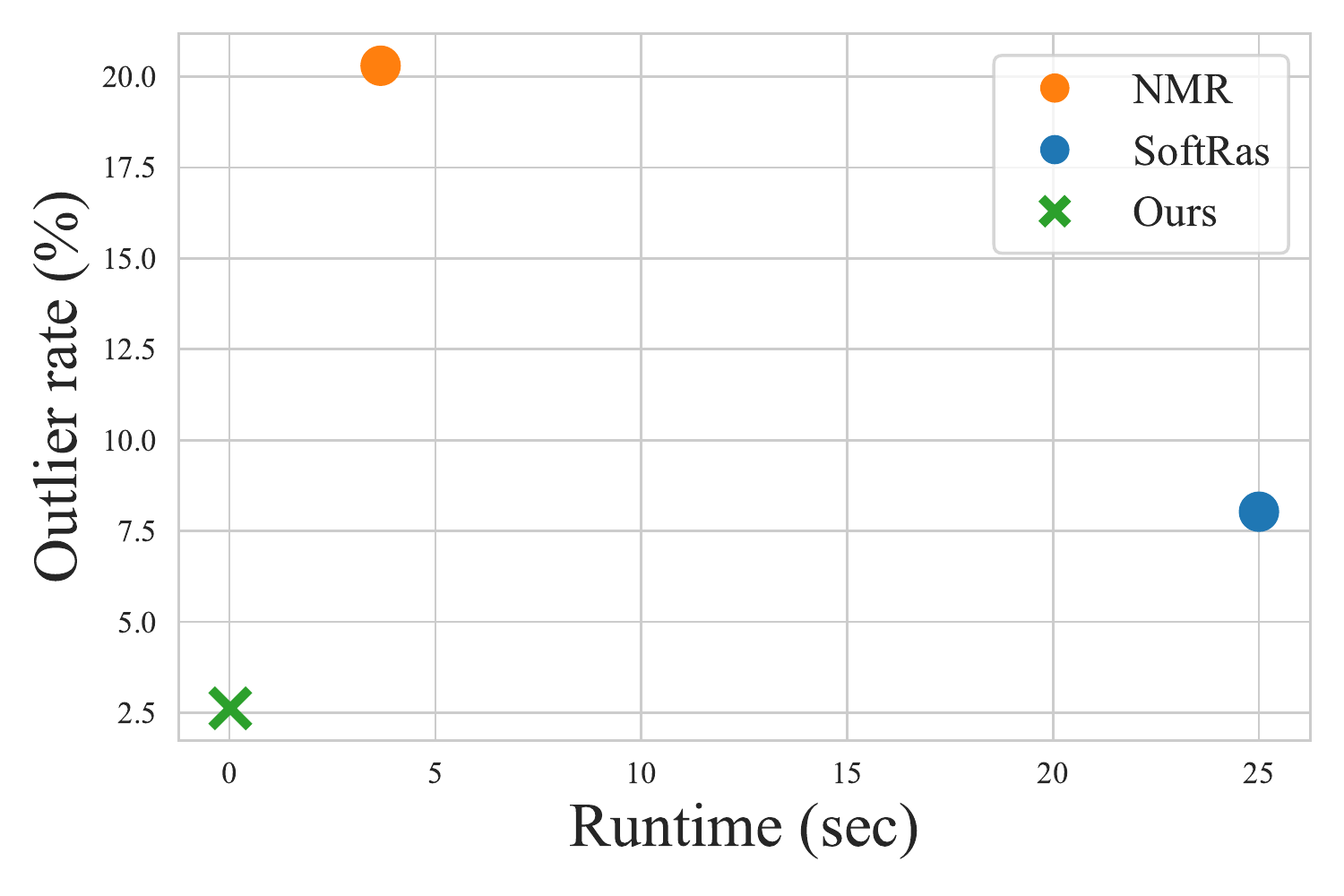}
&\includegraphics[width=0.33\linewidth,trim={1mm 2mm 3mm 0mm},clip]{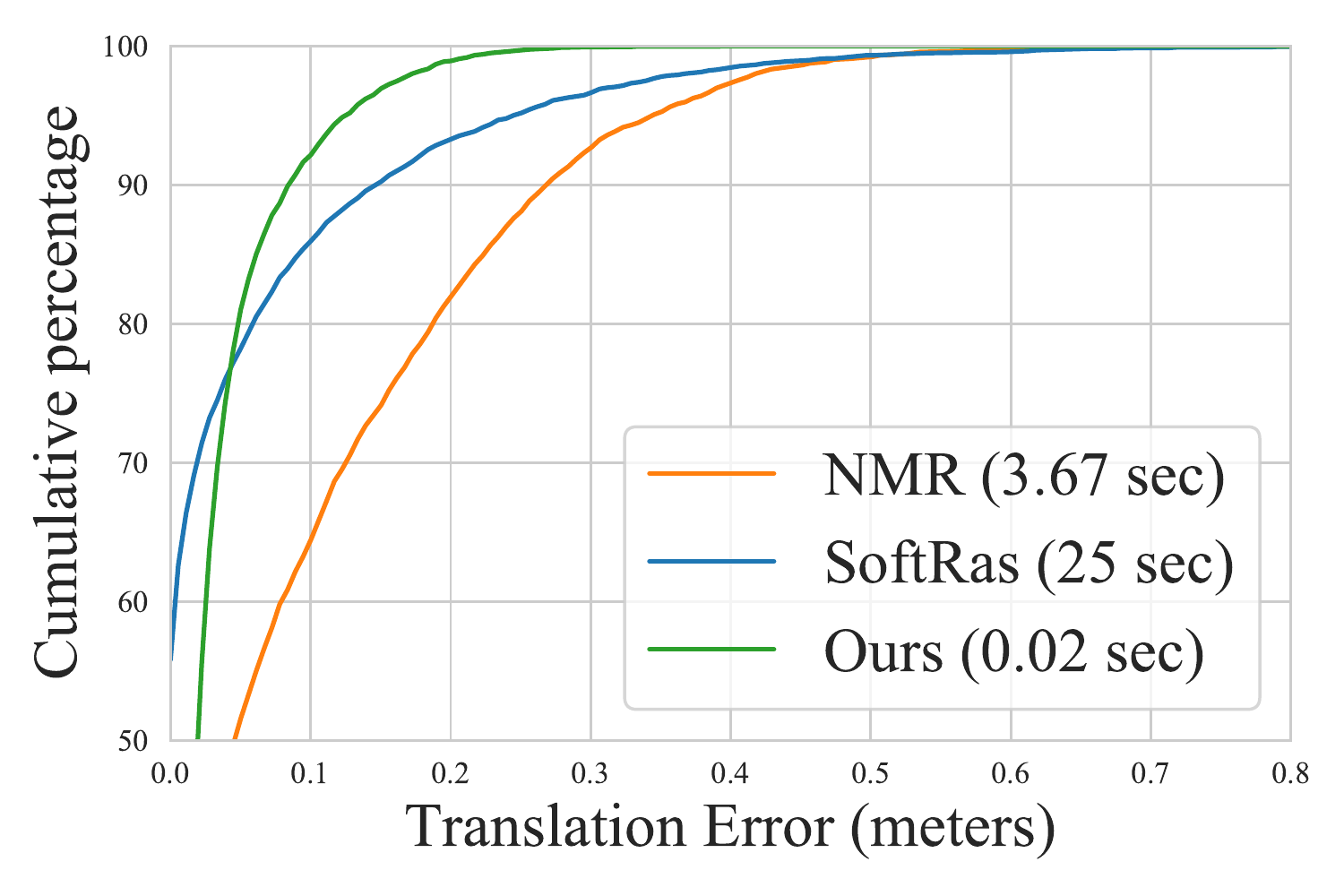}	
&\includegraphics[width=0.33\linewidth,trim={1mm 2mm 3mm 0mm},clip]{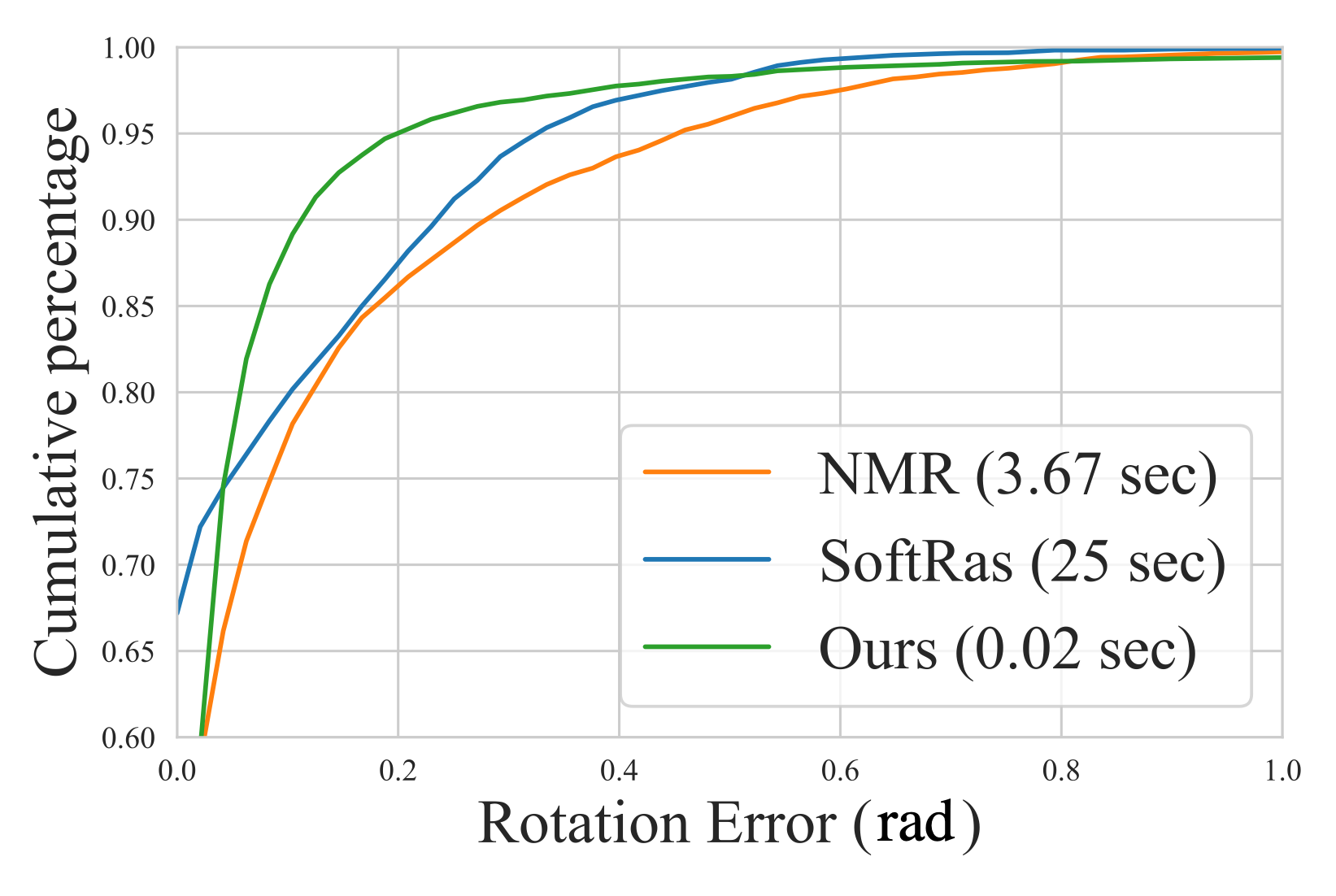}
\\\end{tabular}
}
\arxiv{-3mm}
\caption{
\wcma{
\textbf{Quantitative analysis on 6 DoF pose estimation.} Our deep optimizer is robust, accurate, and significantly faster.
}
}
\label{fig:cum-error}
\arxiv{-3mm}
\end{figure*}

\subsection{Deep Feedback Network}
As we have alluded to above, structured optimization and deep learning have very different yet complementary strengths. Our goal is to bring together the two 
paradigms, and develop a generic approach to inverse problems. 

The key innovation of our approach is  to replace the analytical function $g_E$ defined in structured optimization approach at Eq.~\ref{equ:grad} with a neural network. 
Specifically, we design a neural network $g_\bw$ that takes the  same set of inputs as $g_E$ and outputs the update. The hope is that the model can perceive the difference between the observation $\by$ and the simulated forward results $\by^t$ and then predict a new solution based on the \emph{feedback} signal. %
\wcma{In practice, we employ a simple addition rule and fold the step size, parameter priors all into $g_\bw$}:
\begin{equation}
\label{equ:dso_update}
\bx^{t+1} = \bx^{t} + g_{\bw}(\bx^t, \by^t, \by), \quad \text{where}~\by^t = f(\bx^t).
\end{equation}
The network architectures design depends on the form of observational data $\by$ and solution $\bx$. For instance, for  inverse graphics tasks, we utilize  convolutional neural networks, since the observations are images. 
This not only allows us to sidestep all requirements imposed on $f$ (\eg differentiability), but also removes the need for explicitly defining energies. {Unlike conventional learning based methods, we take both $\by^t$ and $\by$ as input to the update so that we incorporate the feedback signal through comparing the two.}

\wcma{
We derive our final deep structured inverse problem solver by applying the aforementioned update functions in an iterative manner. The algorithm is summarized in Alg.~\ref{alg:dso}.
At each step, the solver takes as input the current solution $\bx^t$, the observation $\by$, and the forward simulated results $\by^t$, and predicts the next best solution as defined in Eq.~\ref{equ:dso_update}. }
{In practice, the stopping criteria could either be based on a predefined iteration number or on checking convergence by measuring the difference between solutions from two consecutive iterations. }

\begin{table}[tb]
\centering
\scalebox{0.95}{
\begin{tabular}{lccccccc}
\specialrule{.2em}{.1em}{.1em}
&\multicolumn{2}{c}{Optimization} &\multicolumn{2}{c}{Trans. Error} &\multicolumn{2}{c}{Rot. Error ($^\circ$)} &Outlier\\
Methods & Step & Time&  Mean & Median &  Mean & Median &  (\%) \\
\hline
NMR \cite{kato2018neural} & 105 & 3.67 s  &0.1 &0.05 &5.78& 1.68  &20.3\\
SoftRas \cite{liu2019soft} & 157 & 25 s &0.05  &0.003  & 4.14 & 0.5 &8.03\\
Deep Regression & 1 & 0.004 s & 0.07 &0.06 & 10.07 &7.68 &  5\\
Ours & 5 & 0.02 s  &0.02 &0.009 & 2.64& 1.02 & 2.6\\
\specialrule{.1em}{.05em}{.05em}
\end{tabular}
}
\caption{\textbf{Quantitative comparison on 6 DoF pose estimation.} %
}
\label{tab:shapenet-quant}
\arxiv{-5mm}
\end{table}

\subsection{Learning}

\wcma{The full deep structured inverse problem solver can be learned in an end-to-end fashion via back-propagation through time (BPTT). 
Yet in practice we find that applying loss function over each stage's intermediate solution $\bx^t$ yields better results. Deep supervision greatly accelerates the speed of convergence.}

\wcma{However, it is non-trivial to design a learning procedure for each iterative update function $g_\bw$, as there exist infinite paths towards the ideal solution. Ideally, we would like our solution to descend towards the ideal solution as quickly as possible. Thus, inspired by \cite{xiong2013supervised}, at each iteration, we learn to aggressively predict the update required to reach the ideal solution. At each stage, the learning procedure finds the best $\bw$ through minimizing the following loss function: }
\[
\arg\min_\bw \sum_{(\by, \bx_\mathrm{gt})}\sum_t \ell(\bx_\mathrm{gt} , \bx^{t} + g_\bw(\bx^t, \by^t, \by)).
\]
\wcma{$\ell$ is a task-specific loss function; for instance, $\ell$ is l2-norm for inverse kinematics.} %

\begin{figure*}[tb]
\centering
\setlength{\tabcolsep}{1pt}
\arxivscale{0.95}{
\begin{tabular}{cccccccccc}
GT & NMR & SoftRas & Regress. & Ours & GT & NMR & SoftRas & Regress. & Ours \\
\includegraphics[width=0.098\linewidth,trim={12mm 2mm 20mm 30mm},clip]{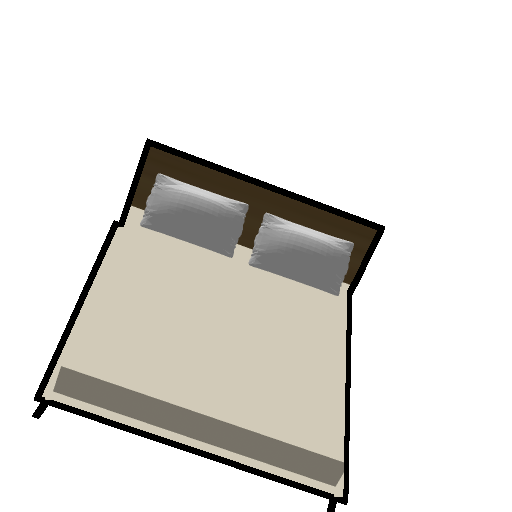} & \includegraphics[width=0.098\linewidth,trim={12mm 2mm 20mm 30mm},clip]{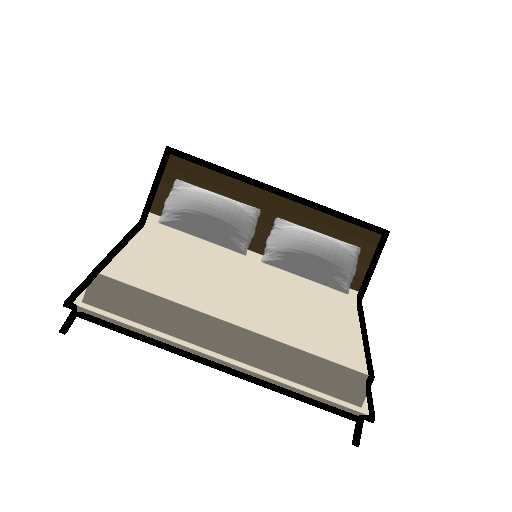} & \includegraphics[width=0.098\linewidth,trim={12mm 2mm 20mm 30mm},clip]{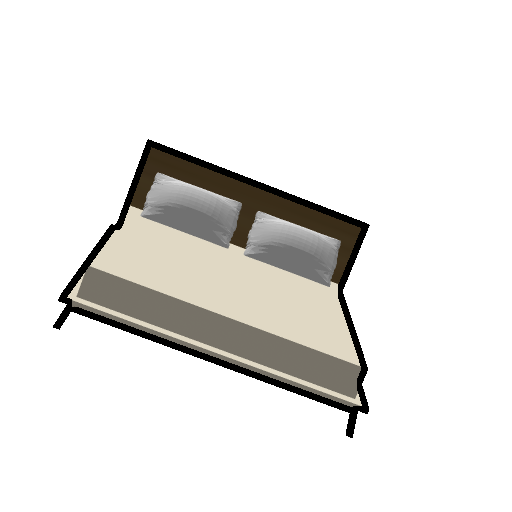} & \includegraphics[width=0.098\linewidth,trim={12mm 2mm 20mm 30mm},clip]{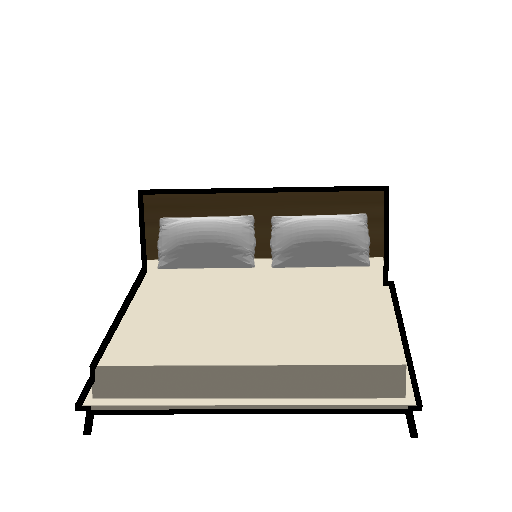} & \includegraphics[width=0.098\linewidth,trim={12mm 2mm 20mm 30mm},clip]{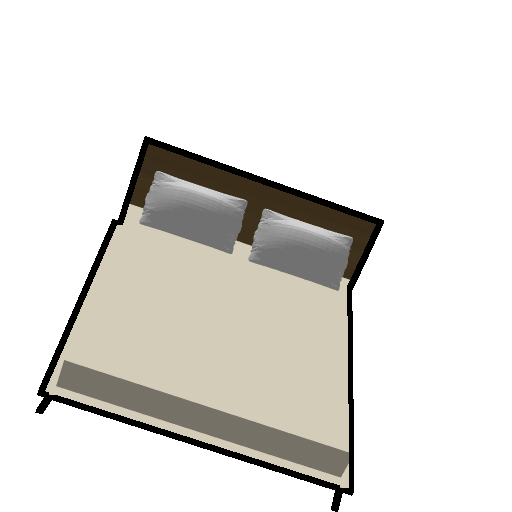}
&\includegraphics[width=0.098\linewidth,trim={12mm 2mm 20mm 30mm},clip]{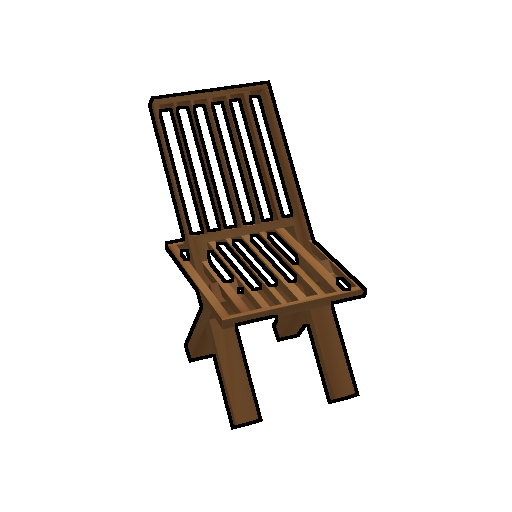} & \includegraphics[width=0.098\linewidth,trim={12mm 2mm 20mm 30mm},clip]{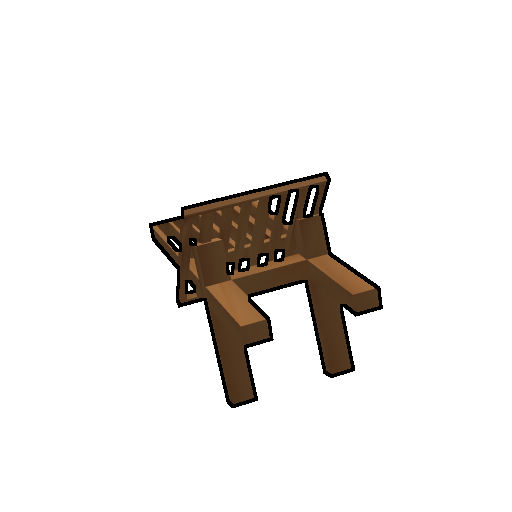} & \includegraphics[width=0.098\linewidth,trim={12mm 2mm 20mm 30mm},clip]{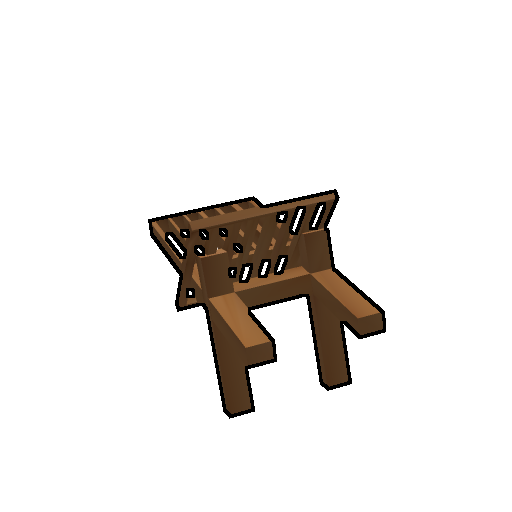} & \includegraphics[width=0.098\linewidth,trim={12mm 2mm 20mm 30mm},clip]{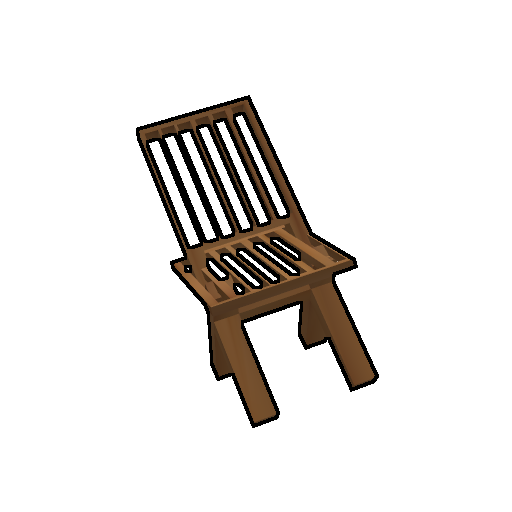} & \includegraphics[width=0.098\linewidth,trim={12mm 2mm 20mm 30mm},clip]{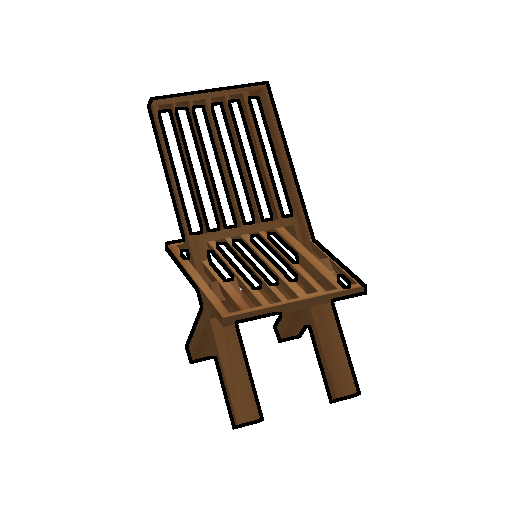} \\
\includegraphics[width=0.098\linewidth,trim={40mm 50mm 40mm 60mm},clip]{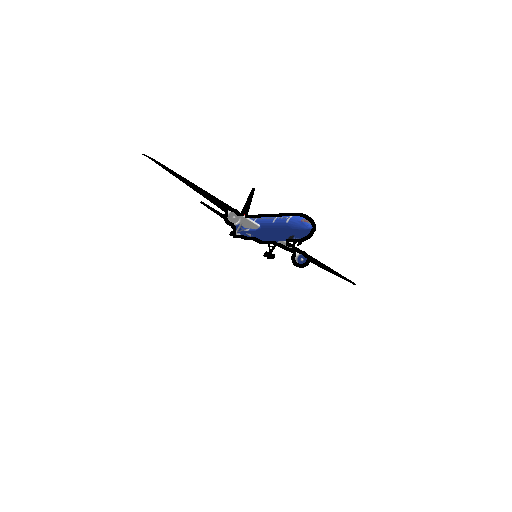} 
& \includegraphics[width=0.098\linewidth,trim={40mm 50mm 40mm 60mm},clip]{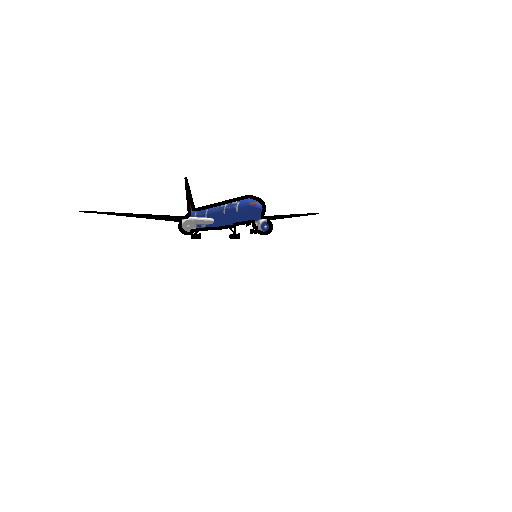} 
& \includegraphics[width=0.098\linewidth,trim={40mm 50mm 40mm 60mm},clip]{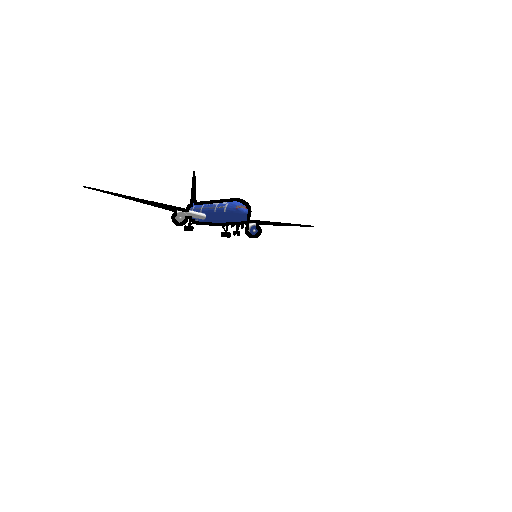} 
& \includegraphics[width=0.098\linewidth,trim={40mm 50mm 40mm 60mm},clip]{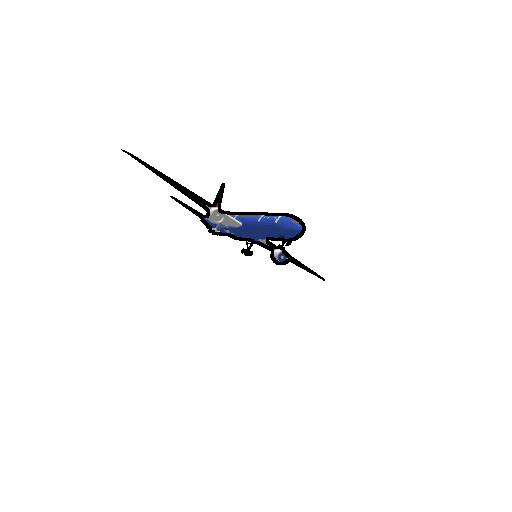} 
& \includegraphics[width=0.098\linewidth,trim={40mm 50mm 40mm 60mm},clip]{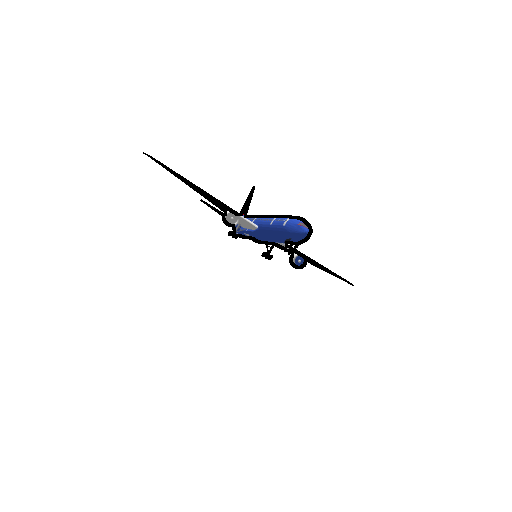}
&\includegraphics[width=0.098\linewidth,trim={12mm 22mm 25mm 40mm},clip]{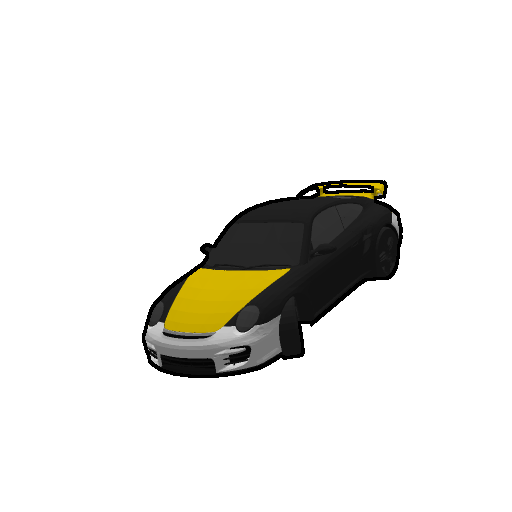} 
& \includegraphics[width=0.098\linewidth,trim={12mm 22mm 25mm 40mm},clip]{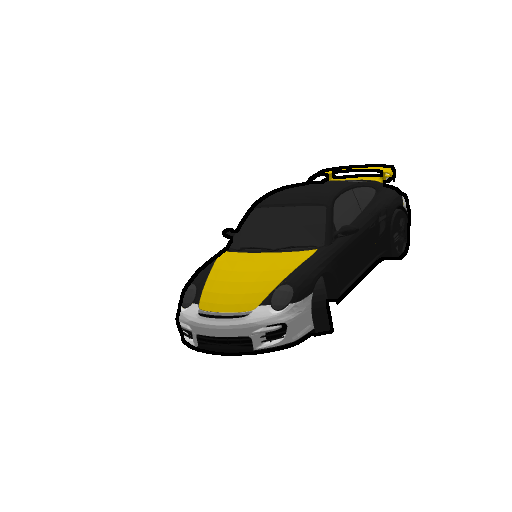} 
& \includegraphics[width=0.098\linewidth,trim={12mm 22mm 25mm 40mm},clip]{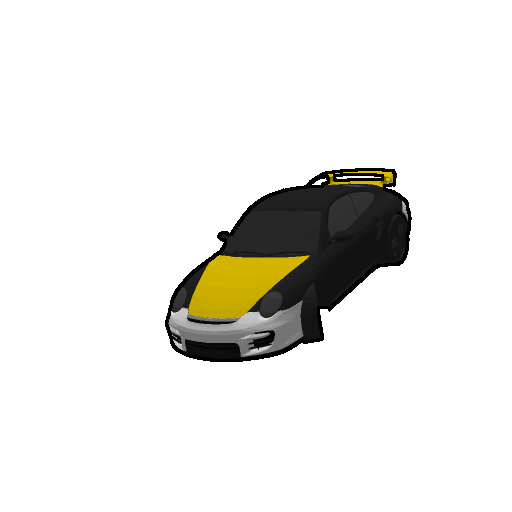} 
& \includegraphics[width=0.098\linewidth,trim={12mm 22mm 25mm 40mm},clip]{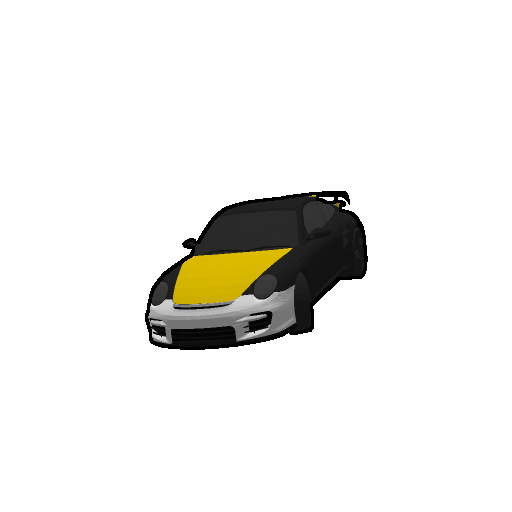} 
& \includegraphics[width=0.098\linewidth,trim={12mm 22mm 25mm 40mm},clip]{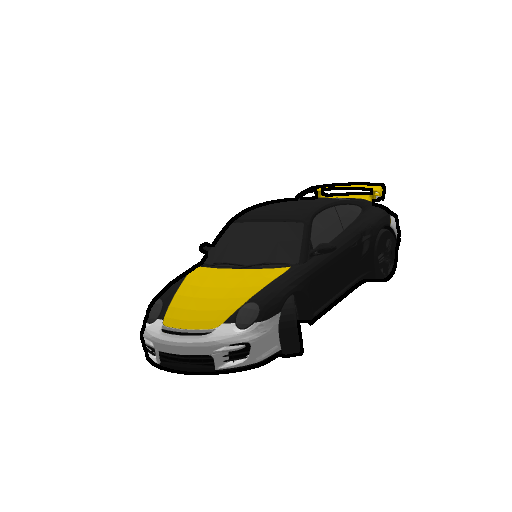} \\
\includegraphics[width=0.098\linewidth,trim={30mm 20mm 15mm 70mm},clip]{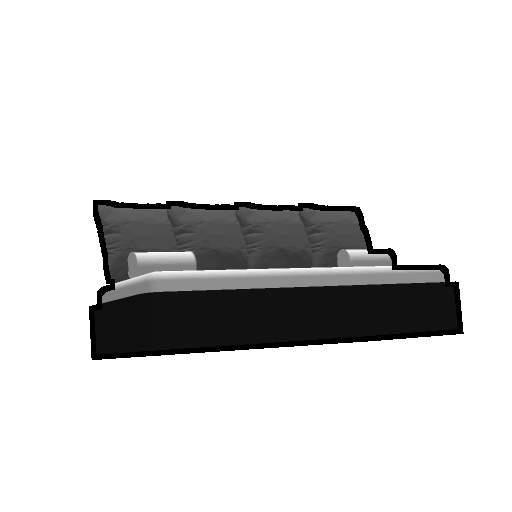} 
& \includegraphics[width=0.098\linewidth,trim={30mm 20mm 15mm 70mm},clip]{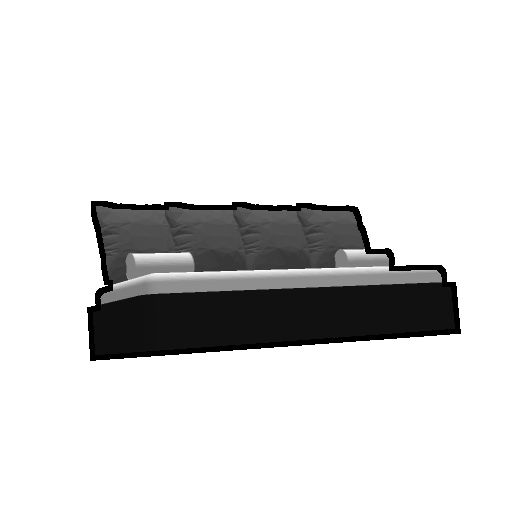} 
& \includegraphics[width=0.098\linewidth,trim={30mm 20mm 15mm 70mm},clip]{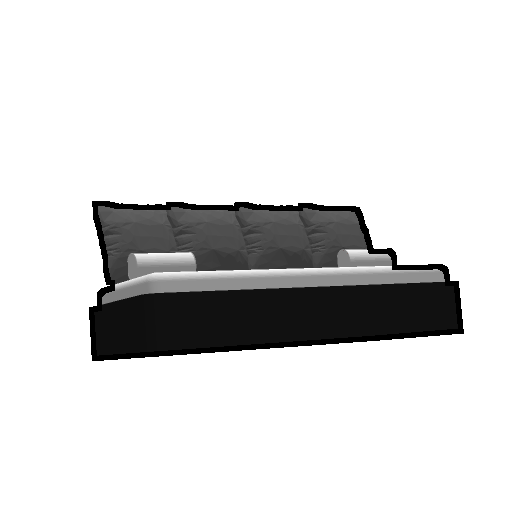} 
& \includegraphics[width=0.098\linewidth,trim={30mm 20mm 15mm 70mm},clip]{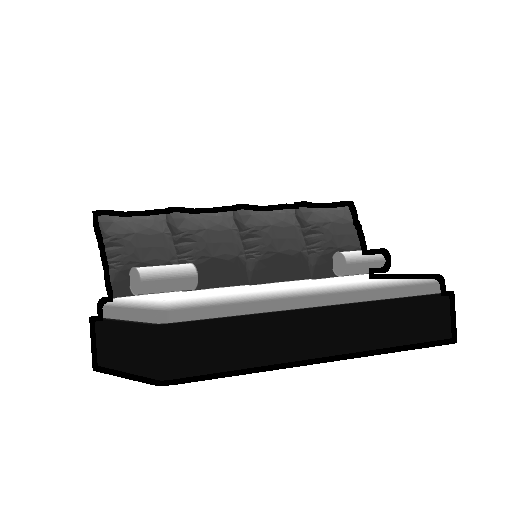} 
& \includegraphics[width=0.098\linewidth,trim={30mm 20mm 15mm 70mm},clip]{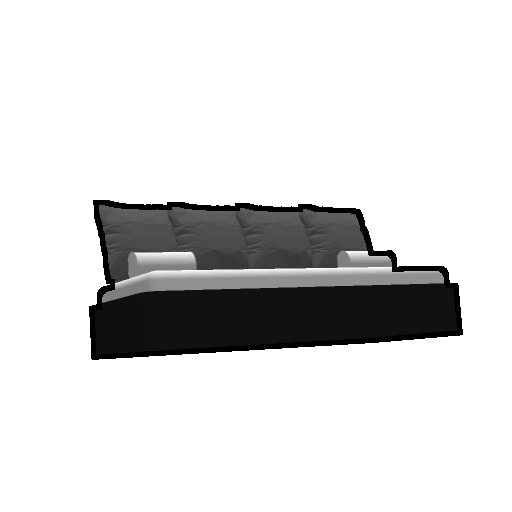}
&\includegraphics[width=0.098\linewidth,trim={12mm 18mm 25mm 35mm},clip]{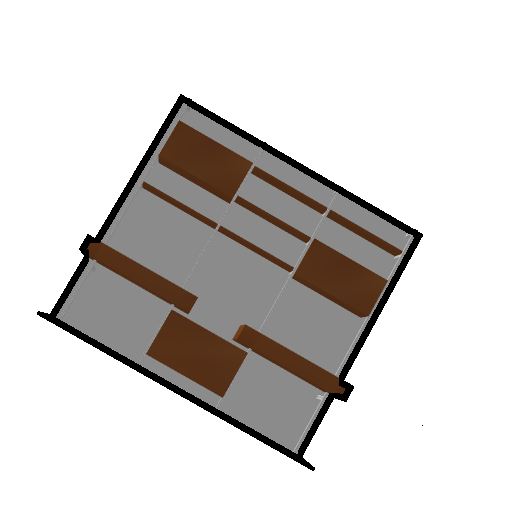} 
& \includegraphics[width=0.098\linewidth,trim={12mm 18mm 25mm 35mm},clip]{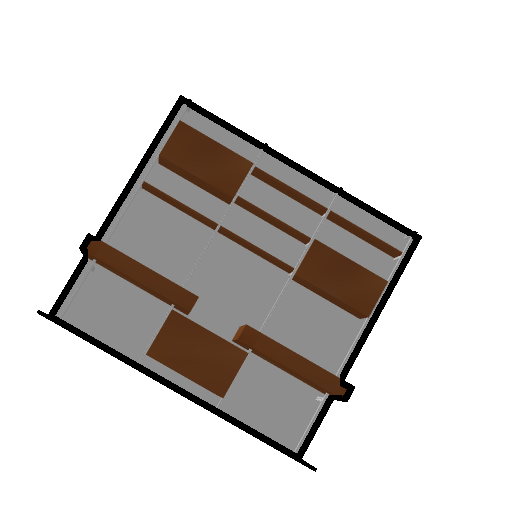} 
& \includegraphics[width=0.098\linewidth,trim={12mm 18mm 25mm 35mm},clip]{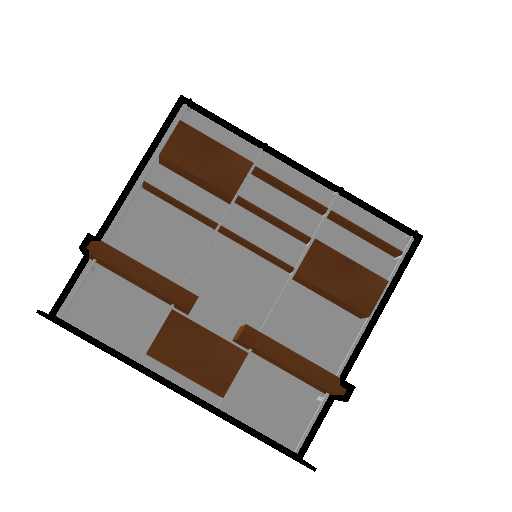} 
& \includegraphics[width=0.098\linewidth,trim={12mm 18mm 25mm 35mm},clip]{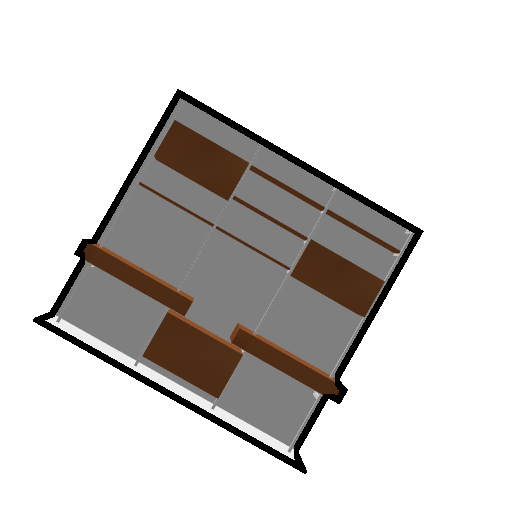} 
& \includegraphics[width=0.098\linewidth,trim={12mm 18mm 25mm 35mm},clip]{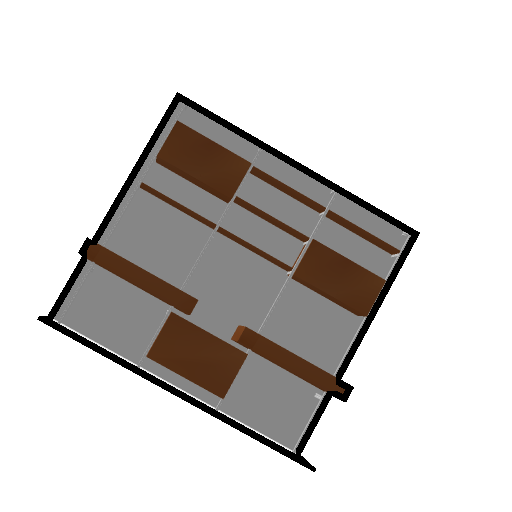} \\
\includegraphics[width=0.098\linewidth,trim={10mm 40mm 20mm 35mm},clip]{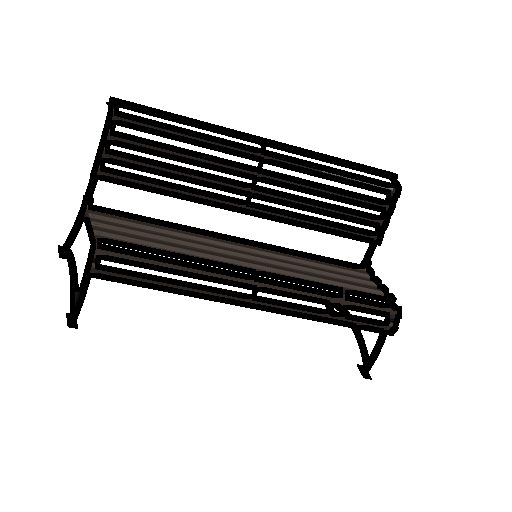} 
& \includegraphics[width=0.098\linewidth,trim={10mm 40mm 20mm 35mm},clip]{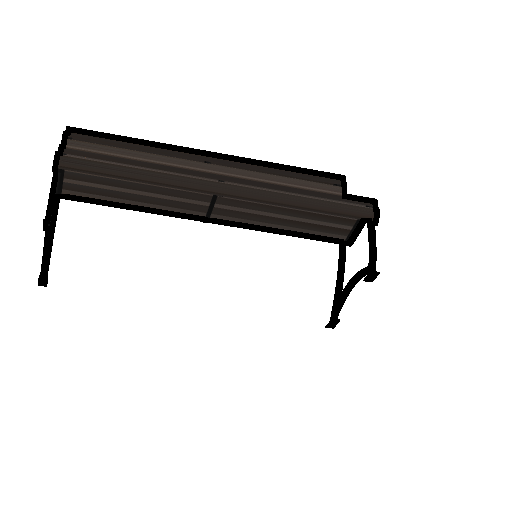} 
& \includegraphics[width=0.098\linewidth,trim={10mm 40mm 20mm 35mm},clip]{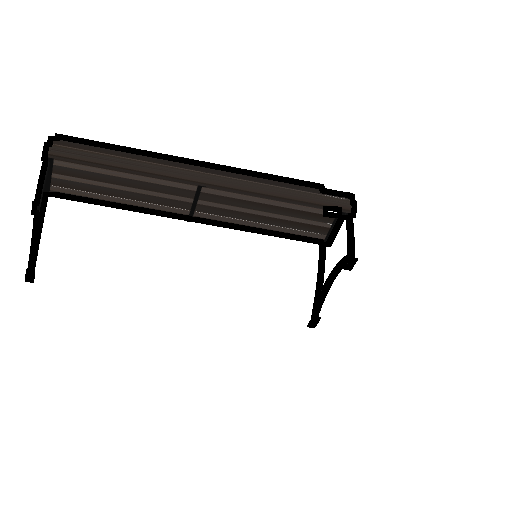} 
& \includegraphics[width=0.098\linewidth,trim={10mm 40mm 20mm 35mm},clip]{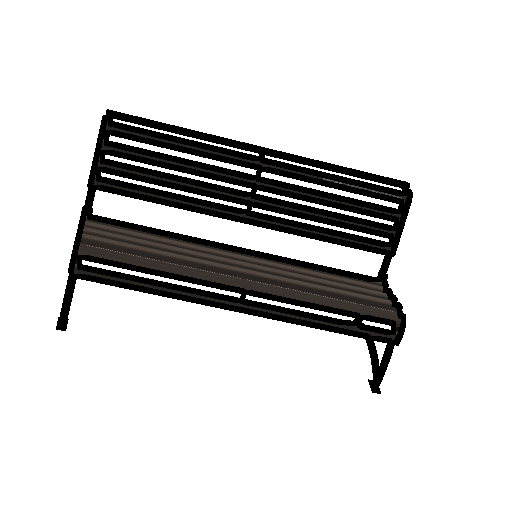} 
& \includegraphics[width=0.098\linewidth,trim={10mm 40mm 20mm 35mm},clip]{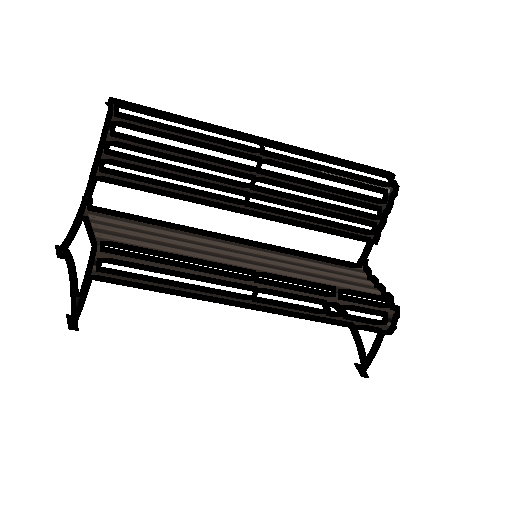}
&\includegraphics[width=0.098\linewidth,trim={12mm 38mm 20mm 35mm},clip]{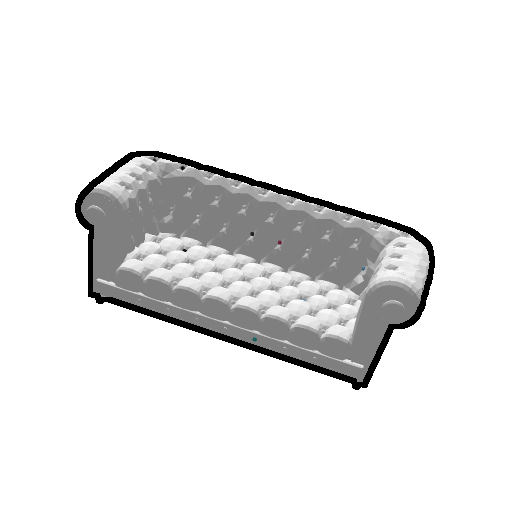} 
& \includegraphics[width=0.098\linewidth,trim={12mm 38mm 20mm 35mm},clip]{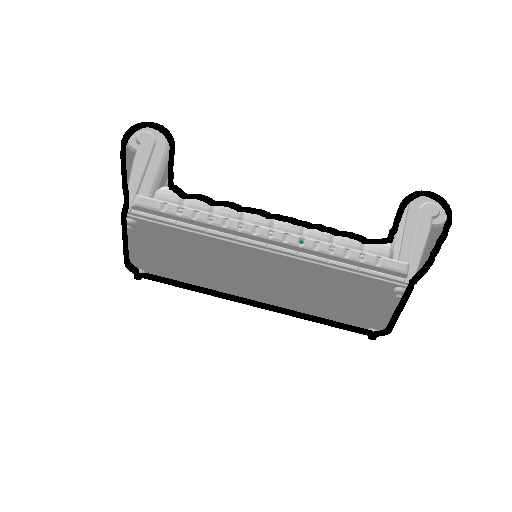} 
& \includegraphics[width=0.098\linewidth,trim={12mm 38mm 20mm 35mm},clip]{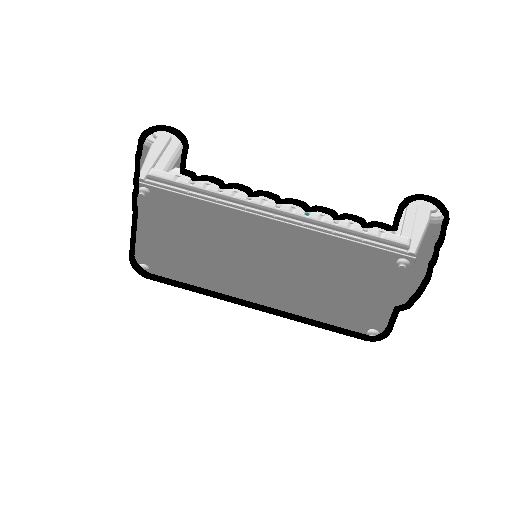} 
& \includegraphics[width=0.098\linewidth,trim={12mm 38mm 20mm 35mm},clip]{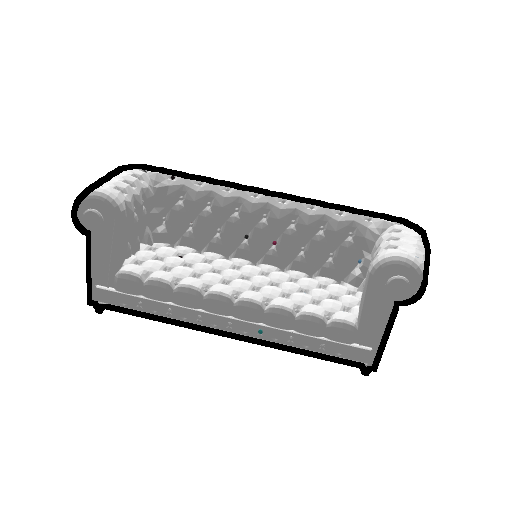} 
& \includegraphics[width=0.098\linewidth,trim={12mm 38mm 20mm 35mm},clip]{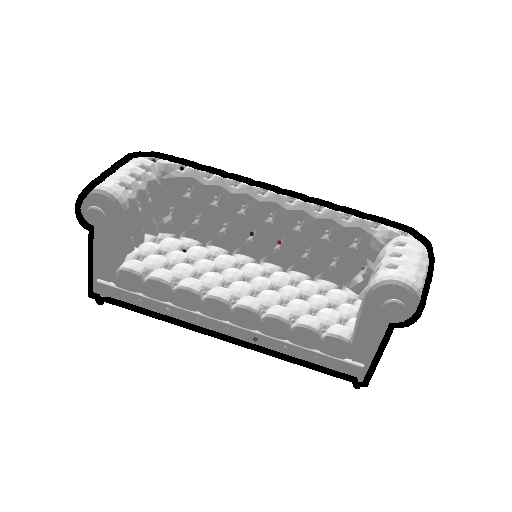} \\
\includegraphics[width=0.098\linewidth,trim={10mm 40mm 12mm 30mm},clip]{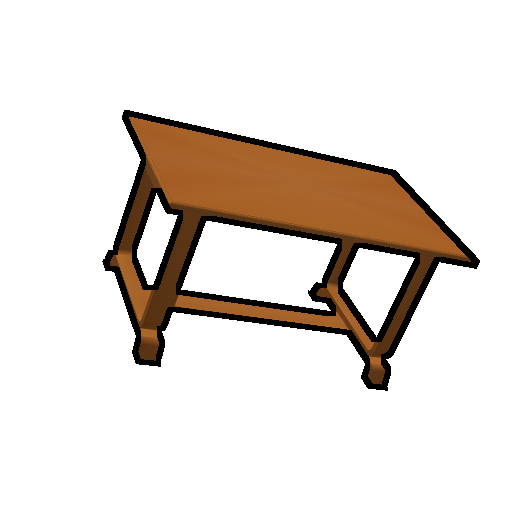} 
& \includegraphics[width=0.098\linewidth,trim={10mm 40mm 12mm 30mm},clip]{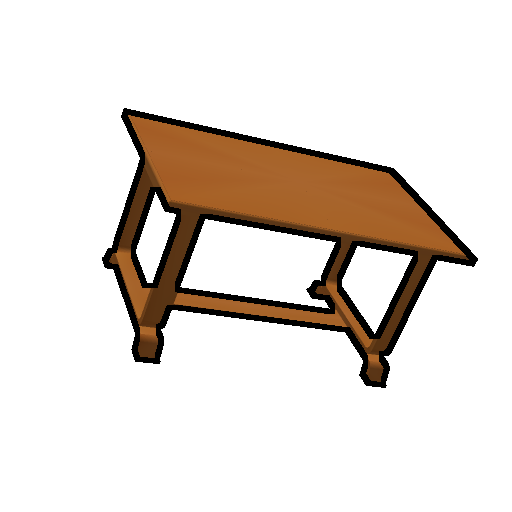} 
& \includegraphics[width=0.098\linewidth,trim={10mm 40mm 12mm 30mm},clip]{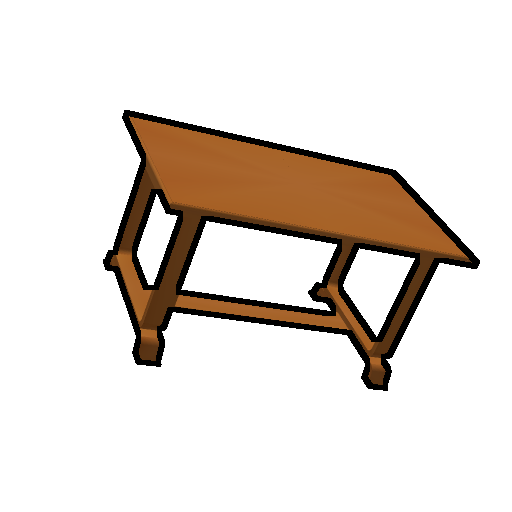} 
& \includegraphics[width=0.098\linewidth,trim={10mm 40mm 12mm 30mm},clip]{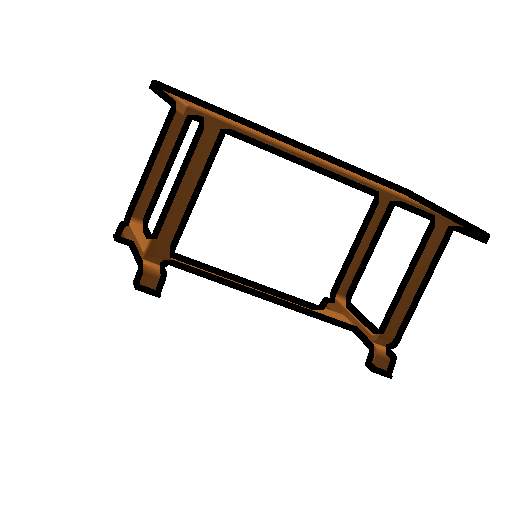} 
& \includegraphics[width=0.098\linewidth,trim={10mm 40mm 12mm 30mm},clip]{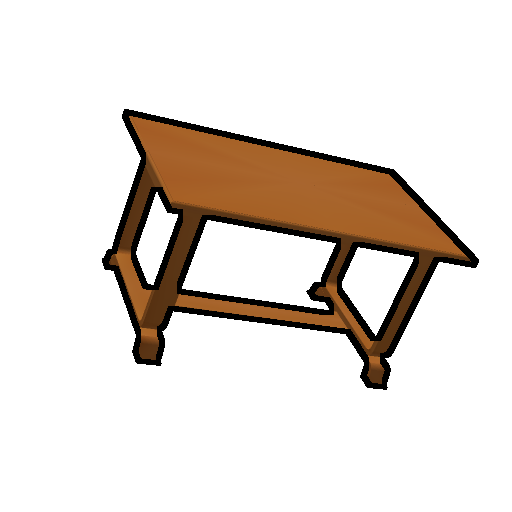}
&\includegraphics[width=0.098\linewidth,trim={12mm 38mm 2mm 25mm},clip]{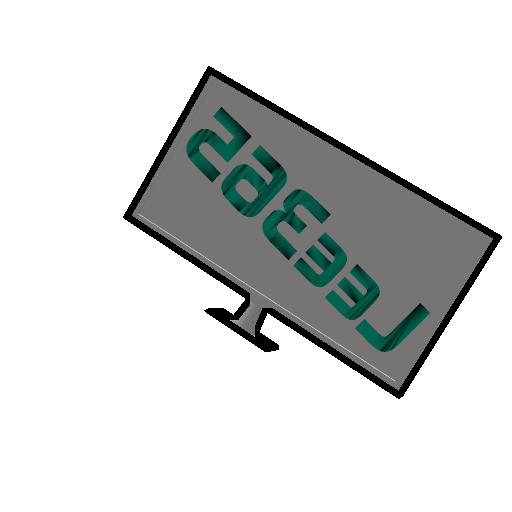} 
& \includegraphics[width=0.098\linewidth,trim={12mm 38mm 2mm 25mm},clip]{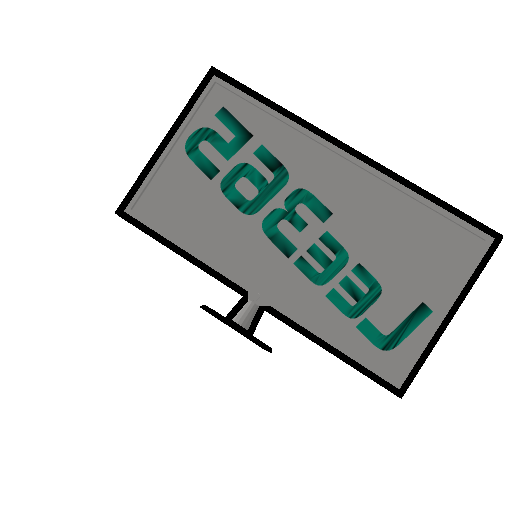} 
& \includegraphics[width=0.098\linewidth,trim={12mm 38mm 2mm 25mm},clip]{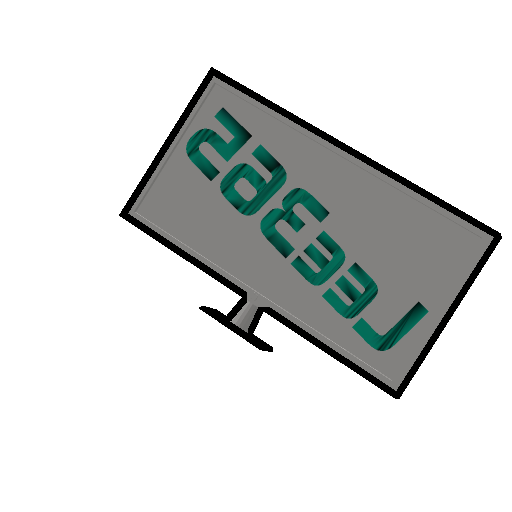} 
& \includegraphics[width=0.098\linewidth,trim={12mm 38mm 2mm 25mm},clip]{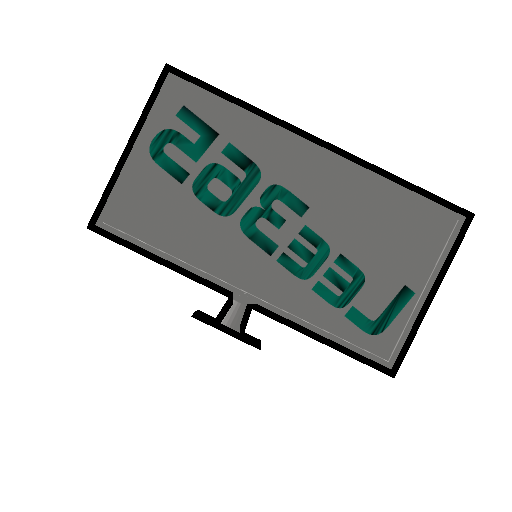} 
& \includegraphics[width=0.098\linewidth,trim={12mm 38mm 2mm 25mm},clip]{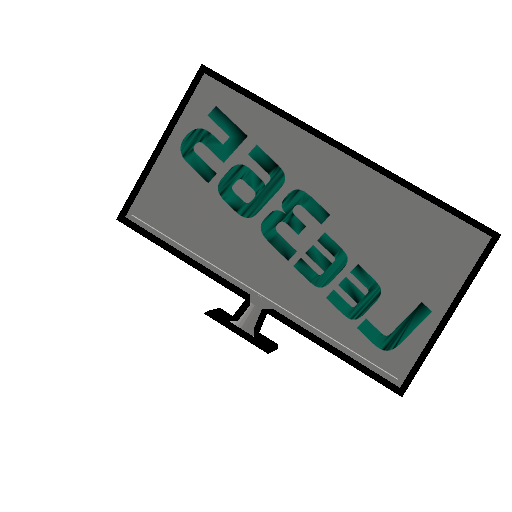} \\

\\\end{tabular}
}
\arxiv{-0.4cm}
\caption{\textbf{Qualitative comparison on 6 DoF pose estimation:}
\wcma{We infer the poses from only silhouette images. The rendered colored images in the figure are for visualization purpose.}
} 
\arxiv{-0.3cm}
\label{fig:qual-comparison-shapenet}
\end{figure*}

\subsection{Discussions}

\paragraph{Stage-wise network:}
{In our standard approach described before, $g_\bw$ is shared across all steps.} 
However, the proximity to the ideal solution varies at different step. {As a consequence, early iteration often takes inputs that are farther to the ideal solution than what a late iteration update step takes. } {This brings difficulties to the network as it needs to handle a variety of output scales across different iteration steps.} %
This motivates us to train a separate update function per step $g_\bw^t(\bx^t, \by^t, \by)$ that better captures the input data distribution at each iteration. To learn this non-shared weight network, we conduct a stage-wise training procedure. We start to train the $g_\bw^0$ first. Then we acquire $\by^0$ for all the training data, which allow us to train $g_\bw^1$. We repeat this procedure until $g_\bw^T$ is trained. {In total $T$ models $\{ g_\bw^t \}$ are trained.} Please refer to the supp. material for the comparison between sharing weights and not sharing weights.

\paragraph{Adaptive update:} Our current update rule is simply an addition, yet it can be easily extended to more sophisticated settings to handle more complex scenarios.
For instance, one can apply the classic momentum technique on top of the predicted gradient to stabilize the optimization trajectory. 
One can also learn another meta-network to dynamically adjust the output of our update network. While all of these options are feasible, we find that in practice a simple strategy suffices. Inspired by the Levenberg-Marquardt method \cite{boyd2004convex}, we exploit a damping factor $\lambda$ to control the effectiveness of the update network, \ie, $\bx^{t+1} = \bx^t + \lambda \cdot g_\bw(\bx^t, \by^t, \by)$. Specifically, $\lambda$ is initialized to 1 at the beginning of each update. If the new estimation results in a lower data energy than that of the original one, we update the estimation. Otherwise we reduce $\lambda$ by half and re-compute. We only need to compute the update gradient once. The forward process is executed on the GPU and hence the computational overhead is negligible.
Through this simple rule, we can guarantee that $E_\mathrm{data}(\bx^t, \by)$ decreases after every iteration. Empirically $\bx^t$ becomes closer to the ground truth $\bx$ as well, since the ambiguity arising from the data term disappears when the estimation is already sufficiently close. \todo{better elaborate what ambiguities; tbh I would remove this sentence, difficult to understand and won't help main story}

\paragraph{Relationship to existing work:} Our model is closely related to the family of iterative networks \cite{greff2016highway,tu2008auto,liang2015recurrent,byeon2015scene,gkioxari2016chained,ramakrishna2014pose,toshev2014deeppose,weiss2010structured,wei2016convolutional,carreira2016human,lin2017inverse,lin2018st}, in particular the
stacked inference machines \cite{ramakrishna2014pose,toshev2014deeppose,weiss2010structured,wei2016convolutional}.
Unlike previous methods that require the model to implicitly learn the relationship between the input and the preceding estimation \cite{oberweger2015training,carreira2016human,wei2016convolutional,zamir2017feedback,manhardt2018deep}, we leverage the forward process to explicitly establish the connection among them and close the loop. This is of crucial importance for inverse problems since the two spaces are very distinct (\eg illumination parameters vs RGB image). The idea of learning to update is inspired by supervised descent methods \cite{xiong2013supervised}. However, unlike their approach we learn the mapping and the feature simultaneously. Furthermore, we focus on inverse problems and design a closed-loop scheme to incorporate feedback signals, while they simply perform iterative update in an open loop setting. Developed independently, Flynn \etal \cite{flynn2019deepview} propose a similar approach for view synthesis. Their model, however, relies on the analytical gradient components. They thus requires the system to be differentiable.
In contrast, our approach directly predicts the update from the observation and the feedback signal. We do not require explicit gradient computation and do not have such a limitation. \addon{Similar to our work, LiDO \cite{LiDO} also leverages deep networks to optimize the latent parameters. Yet unlike their model which directly regresses the GT, we predict the residual instead. This is very critical as the magnitude of the update is strongly correlated with the input difference. The larger (smaller) the input difference is, the more (less) update is required. Predicting the residual can thus significantly ease the learning process. Furthermore, while their prediction networks share weights across all iterations, our feedback networks differ at each step. 
We are able to model the variety of output scales more effectively.
Our work also shares similar insights as \cite{oberweger2015training,carreira2016human,li2018deepim} with a few key differences: (1) rather than relying on the network to \emph{implicitly} establish the relationships between the feedback signal and the observation, we \emph{explicitly} consider the difference and predict an update based upon it. (2) Our model is motivated by classic optimization approaches. We borrow ideas from traditional literature to improve the performance (\ie adaptive update), whereas \cite{li2018deepim} simply unrolls the network, \cite{carreira2016human} employs a bounded, fixed update, and \cite{oberweger2015training} encourages the update network to improve the estimation, no matter what scale it is.
(3) We present a generic framework that is applicable to a wide range of inverse problems, while \cite{oberweger2015training,carreira2016human,li2018deepim} are specialized to respective specific task.}
{We refer the readers to the supp. material for more detailed discussions on reinforcement learning and prior art \cite{carreira2016human,li2018deepim,LiDO,oberweger2015training}.}

\paragraph{Applicability:} \wcma{Unlike previous work, our approach neither has restrictions on the forward process $f$, nor need to construct domain-specific objectives at test time. 
During inference, at each iteration, we simply adopt a feed-forward operation $g$ on top of current estimate and predict the update.
Our method is applicable to a wide range of tasks so long as the forward process function $f$ is available. In the following sections, we showcase our approach on two different inverse graphics tasks (object pose estimation and illumination estimation from a single image) as well as one robotics task (inverse kinematics).}

%
%

%
%
%
%

%
%
%
%
%
%
%
%
%
%
%
%

%

%

%
%

%
%

%

%
%

%

%

%

%

%
%

%
%

%
%

%
%
%
%
%
%
%
%

%
%

%
%
%
%
%
%

%

%

%

%

%
%

%
%
%
%

%
%
%
%
%
%
%

%

%

%

%

%

%

%

%

%
%
%
%

%

%
%
%
%
%
%

%
%
%
%
%
%
%

%

%
%
%
%
%
%
%
%
%
%
%
%
%

%

%

%

%

%
%
%

%
%

%
%
%
%
%
%
%
%

%

%
%

%

%

%
%
%

%
%
%
%
%
%
%
%

%

%
%

%
%
%

%
%
%
%
%
%
%
%

%
%
%

%
%

%
%

%

%
%
%
%
%
%
%
%
%

%
%
%
%
%
%
%
%
%
%
%
%
%
%
%
%
%
%

%
%
%

\begin{figure}[t]
\begin{minipage}[c]{0.41\textwidth}
\resizebox{0.95\textwidth}{!}{%
\setlength{\tabcolsep}{3pt}
\begin{tabular}{lccc}
\specialrule{.2em}{.1em}{.1em}
& Forward & Inverse\\
Module & Rendering & Update & Total\\
\hline
NMR \cite{kato2018neural} & 28 ms & 7 ms & 35 ms\\
SoftRas \cite{liu2019soft} & 76 ms & 84 ms & 160 ms\\
Ours & 2.6 ms & 0.9 ms & 3.5 ms\\
\specialrule{.1em}{.05em}{.05em}
\end{tabular}
}
\captionof{table}{\textbf{Runtime breakdown of a single optimization step for 6 DoF pose estimation.}}
\label{tab:pose-runtime}
\end{minipage}
\hfill
\begin{minipage}[c]{0.57\textwidth}
	\resizebox{\textwidth}{!}{%
	\setlength{\tabcolsep}{1pt}
	\begin{tabular}{cc}
	\includegraphics[width=0.5\linewidth,trim={4mm 0mm 3mm 3mm},clip]{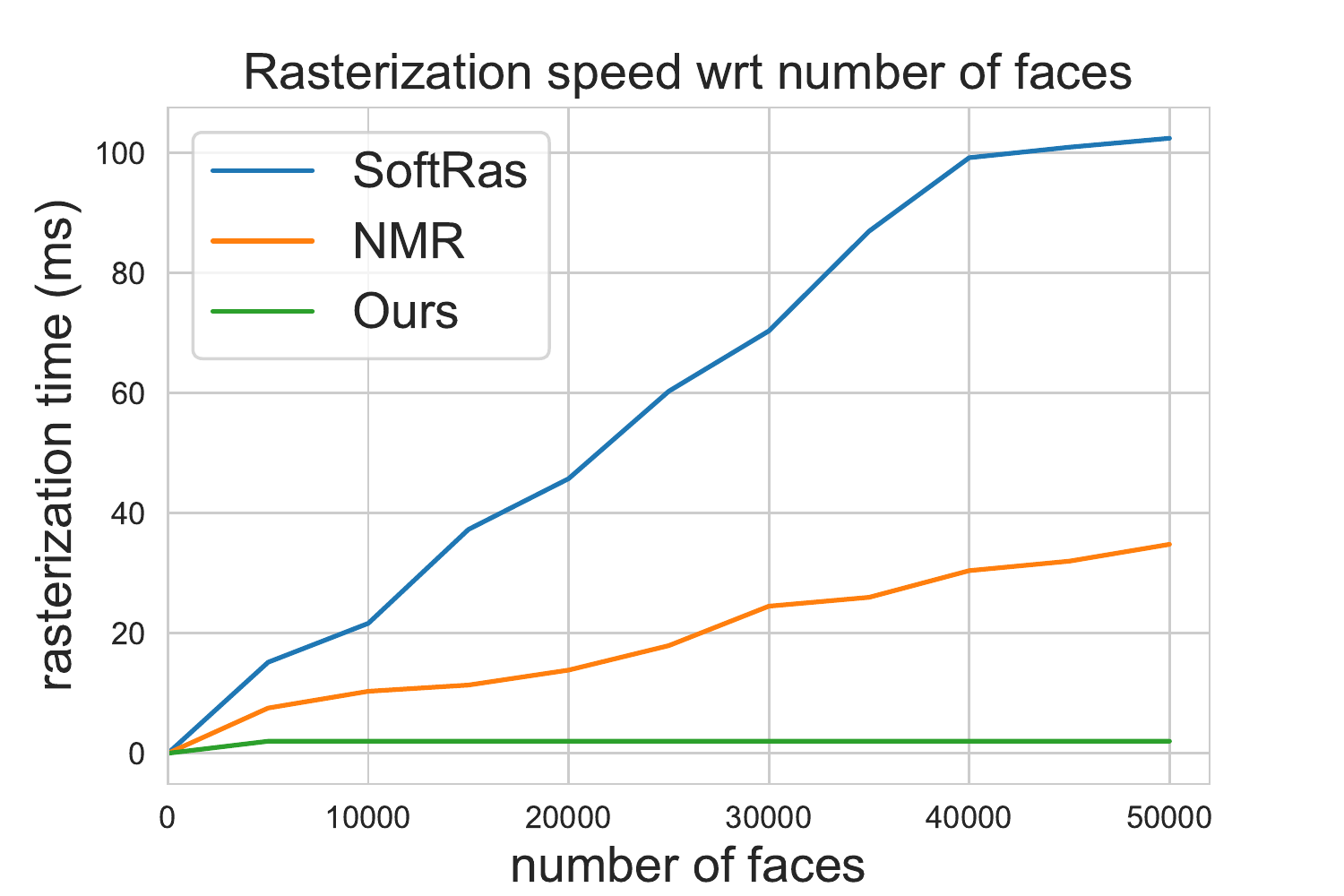}
	&\includegraphics[width=0.5\linewidth,trim={4mm 2mm 3mm 3mm},clip]{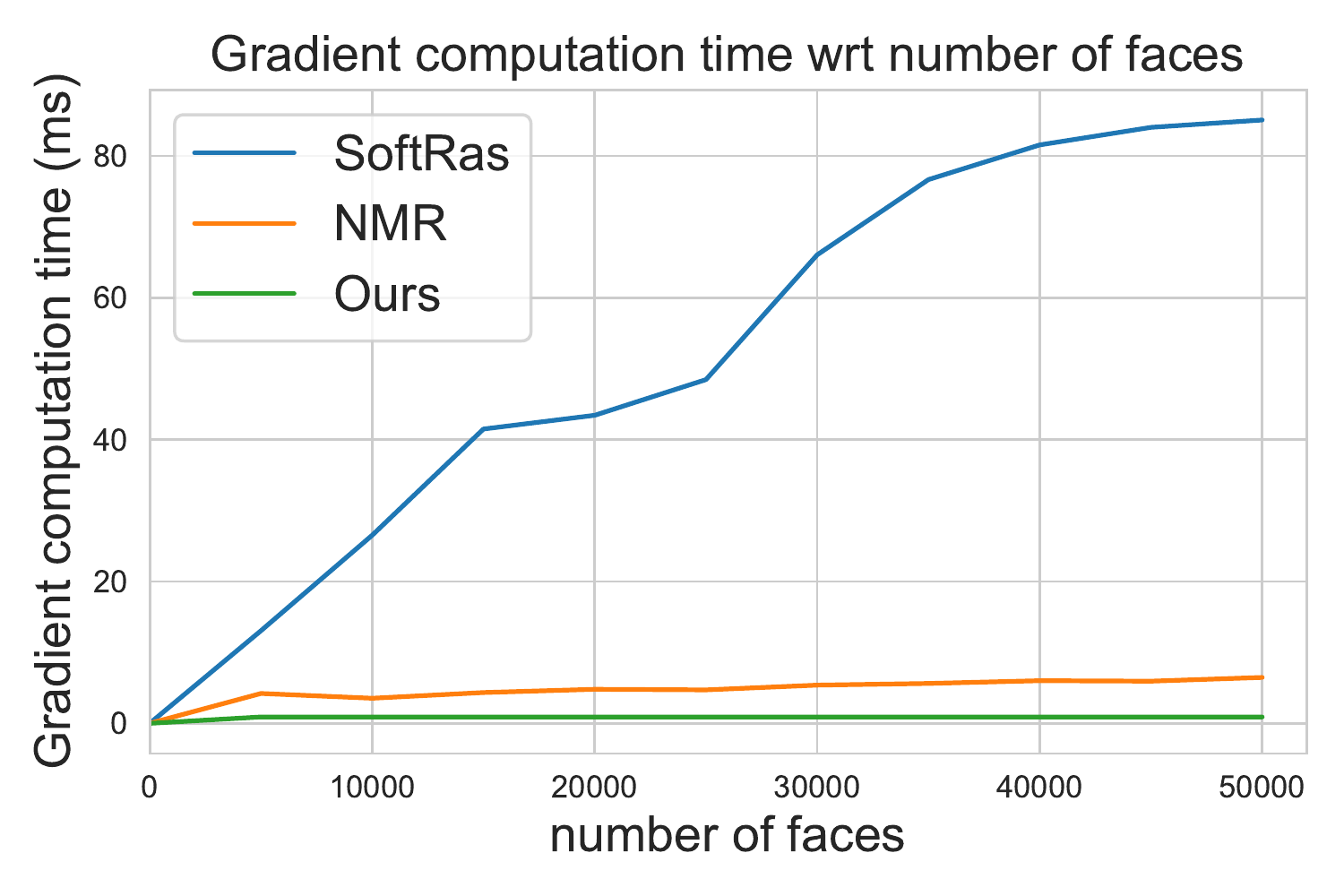}
	\\\end{tabular}
	}
    \caption{\textbf{Runtime vs number of faces.} (Left) Forward rasterization time. (Right) Backward gradient computation (inverse update) time.}
	\label{fig:runtime-faces-num}

\end{minipage}
\end{figure}

\section{Application I: 6-DoF Object Pose Estimation}
\label{sec:pose-estimation}
\arxiv{-3mm}

\paragraph{Problem formulation:} 
Assume that the 3D model of the object is given \cite{hinterstoisser2012model,cao2016real} and the camera intrinsic parameters are known. For {a given object pose  wrt the camera, denoted as} $\bx \in$ SE(3), we can generate the corresponding image observation $\by$ through a forward rendering function $f: \bx \rightarrow \by${, powered by a graphics engine}. The goal of 6 DoF pose estimation is to \emph{invert} the process and recover the latent pose $\bx$ from the observation image $\by$. {This problem is particularly important for problems such as robot grasping \cite{li2018deepim} and self-driving \cite{ma2019deep}.}
Unlike previous approaches that leverage RGB information or depth geometry to guide the pose estimation, we focus on a more challenging setting where the observation is \emph{a single silhouette image} $\by \in \{0, 1\}^{H\times W}$. %
{The object pose $\bx = (\bx_\mathrm{quat}; \bx_\mathrm{trans})$ is represented by a unit quaternion for rotation $\bx_\mathrm{quat}$ and a 3D translation vector $\bx_\mathrm{trans}$. } 

{\paragraph{Data:} We use the 3D CAD models from ShapeNet \cite{chang2015shapenet} within 10 categories: cars, planes, chairs, bench, table, sofa, cabinet, bed, monitor, and couch. The dataset is split into training ($70\%$), validation ($10\%$) and testing ($20\%$). 
For each object, we randomly sample an axis from the unit sphere and rotate the object around the axis by $\theta \sim [-40, 40]$ degrees. 
We further translate the object along each axis by a random offset within $[-0.2, 0.2]$ meters.
Given the randomly generated ground truth object poses, we render $128 \times 128$ silhouette images with non-differentiable PyRender \cite{pyrender} as input observations. 
We refer the readers to the supp. material for the performance of our model on other image sizes.}

\begin{table}[tb]
\centering
\scalebox{0.95}{
\begin{tabular}{lcccc}
\specialrule{.2em}{.1em}{.1em}
Training on $0^\circ-40^\circ$& \multicolumn{2}{c}{Trans. Error} & \multicolumn{2}{c}{Rot. Error ($^\circ$)}\\
Evaluation Rot. Range & Mean & Median & Mean & Median\\
\hline
$40^\circ-45^\circ$ & 0.05& 0.03& 11.33 & 4.97 \\
$45^\circ-50^\circ$ & 0.05& 0.04& 15.62 & 5.60\\
$50^\circ-55^\circ$ & 0.06& 0.04 & 18.58 & 6.86\\
$55^\circ-60^\circ$ & 0.07& 0.05 & 24.14 & 9.58\\
\specialrule{.1em}{.05em}{.05em}
\end{tabular}
}
\caption{\textbf{Test on unseen rotations.}}
\label{tab:shapenet-unseen}
\arxiv{-3mm}
\end{table}

\paragraph{Metrics:} We measure the translation error with euclidean distance and the rotation error with angular difference. Inspired by \cite{Geiger2012CVPR}, we also compute the \emph{outlier ratio} as an indicator of the general quality of the output. Specifically, we define the prediction to be an outlier if the translation error is higher than 0.2 or the rotation error is larger than 30$^{\circ}$.

\paragraph{Network architecture:} 
Our deep feedback network $g_\bw$ is akin to the classic LeNet \cite{lecun1998gradient}. It takes as input the rendered image ${\by^t} = f(\bx^t)$, the observed image $\by$, as well as the difference image $\hat{\by} - \by^t$, and directly outputs the update $\Delta\bx$. We apply an additional normalization operator over the rotation component to correct it to a valid unit quaternion. We unroll our deep feedback network for five steps.
MSE is employed as the loss function for both rotation and translation since it produces the most stable results. 

\paragraph{Baselines:} For optimization methods, the energy function consists of a data term $E_\mathrm{data}(f(\bx), \by)$ that favors agreement and a prior term $E_\mathrm{prior}(\bx)$ that encourages the quaternion to remain on the manifold.
To make the forward rendering procedure $f$ differentiable, we utilize the state-of-the-art differentiable renderers for comparison, \ie neural mesh renderer (NMR\cite{kato2018neural}) and soft rasterization (SoftRas \cite{liu2019soft}). 
We utilize the following stopping criteria for the optimizer: (i) 500 iterations, or (ii) the silhouette difference between the observation and the one generated by the renderer stops improving for 20 iterations. 
For the deep regression method, we use the same architecture as our deep feedback network except that no feedback is provided.

\paragraph{Results:} 
As shown in Tab. \ref{tab:shapenet-quant}, our method achieves a significantly lower outlier ratio compared to other approaches. This indicates that our model is more robust and less susceptible to becoming stuck in local optimum. It also has comparable performance to differentiable renderers in terms of mean translation and angular error, while being two to three orders of magnitude faster. On the other hand, our method has much better performance than the non-feedback deep regression method.  %
For the category-wise performance, please refer to the supp. material. 
Fig.~\ref{fig:qual-comparison-shapenet} showcases some qualitative results. Our method is robust to extreme poses, whereas optimization based method is easy to get stuck in a local optimum. 
\

\begin{table}[tb]
\centering
\scalebox{0.9}{
\begin{tabular}{lcccccccc}
\specialrule{.2em}{.1em}{.1em}
&\multicolumn{2}{c}{Optimization} &\multicolumn{3}{c}{Directional light} &\multicolumn{3}{c}{Point light} \\
Methods & Step & Time&  Mean & Median & Outliers &  Mean & Median & Outliers \\
\hline
NMR\footnote{NMR does not support point light. Furthermore, its directional light is highly simplified and did not consider self-occlusion. }\cite{kato2018neural} &166.7  &  58.3 s  & 0.099 & 0.037 & 19.2\% & - & - & -\\
Deep regression \cite{janner2017self} &1  &0.043 s  &  0.067 & 0.022& 24\%  & 0.111& 0.084 & 11\% \\
Ours &7  &0.183 s  & 0.052 & 0.008 & 8\%  & 0.084& 0.064& 9\% \\
\specialrule{.1em}{.05em}{.05em}
\end{tabular}
}
\caption{\textbf{Illumination estimation on ShapeNet.}}
\label{tab:light-quant}
\arxiv{-4mm}
\end{table}

\paragraph{Deep feedback network as initialization:} Due to the highly non-convex structure of the energy model, a good initialization is required for optimization methods to achieve good performance. 
One natural solution is to exploit our model as an initialization and employ classic solvers for the final optimization. By combining our approach with SoftRas, we can further reduce the error by more than $50\%$.
We refer the readers to supp. material for detailed analysis.

\paragraph{Runtime analysis:} \wcma{We show the runtime break down for a single update step in Tab.~\ref{tab:pose-runtime} and the runtime w.r.t the number of faces in Fig. \ref{fig:runtime-faces-num}.} As we neither need to construct the computation graph nor storing any activation value for gradient computation during the forward rasterization process, our rendering is significantly faster.  
For gradient computation, SoftRas is far slower as it needs to propagate the gradient to multiple faces. In contrast, our update model is simply an efficient feed-forward neural net that takes as input the (difference) silhouette images. Its speed is invariant to the number of faces. 

\begin{figure*}[tb]
\centering
\setlength{\tabcolsep}{1pt}
\arxivscale{0.98}{
\begin{tabular}{cccccccc}
GT & NMR & Regress. & Ours & GT & NMR & Regress. & Ours\\
\includegraphics[width=0.12\linewidth,trim={25mm 32mm 15mm 13mm},clip]{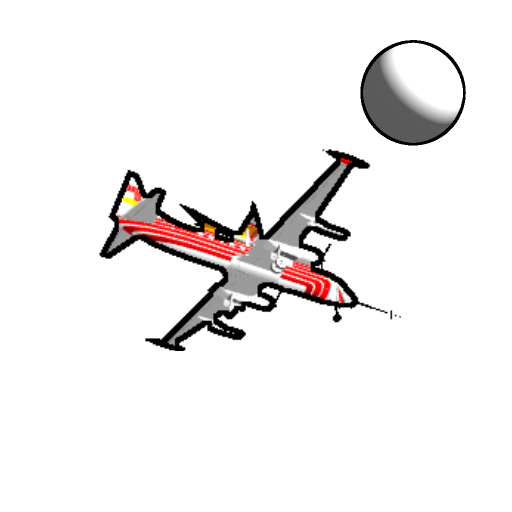}  
& \includegraphics[width=0.12\linewidth,trim={25mm 32mm 15mm 13mm},clip]{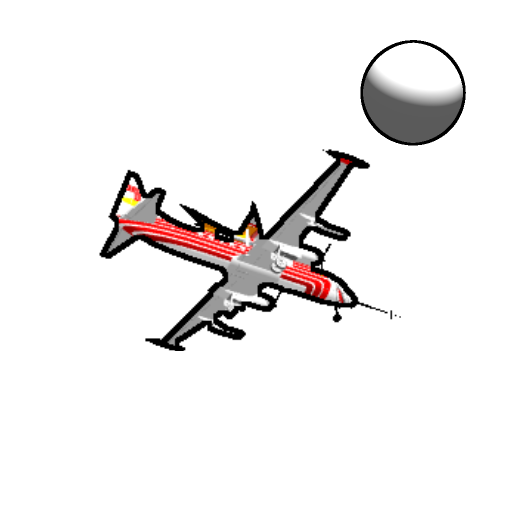}  
& \includegraphics[width=0.12\linewidth,trim={25mm 32mm 15mm 13mm},clip]{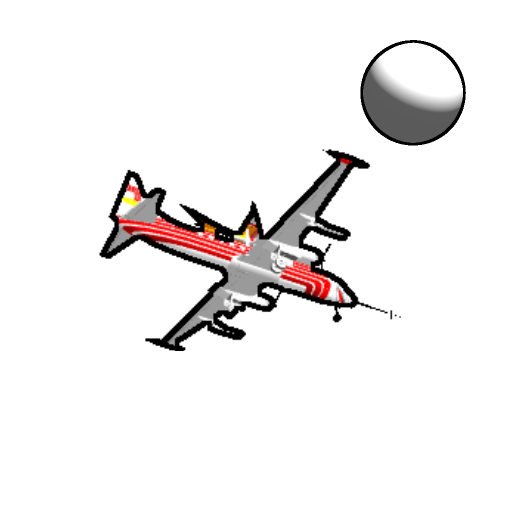}
& \includegraphics[width=0.12\linewidth,trim={25mm 32mm 15mm 13mm},clip]{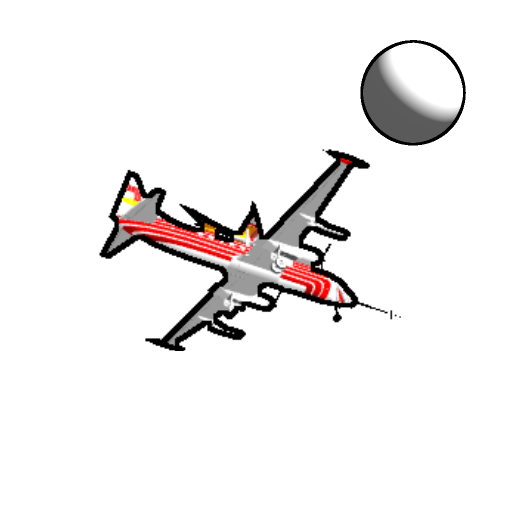}
&\includegraphics[width=0.12\linewidth,trim={25mm 32mm 15mm 13mm},clip]{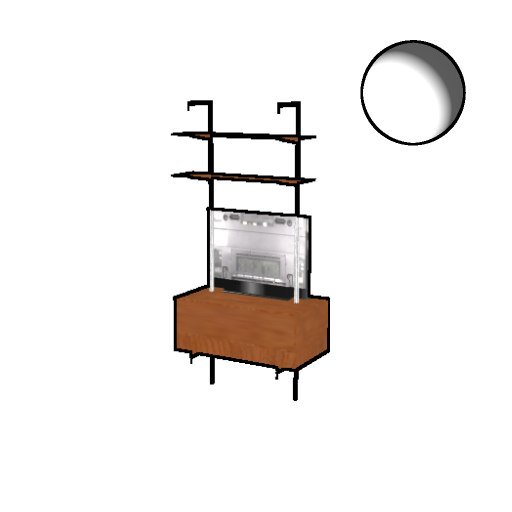}  
& \includegraphics[width=0.12\linewidth,trim={25mm 32mm 15mm 13mm},clip]{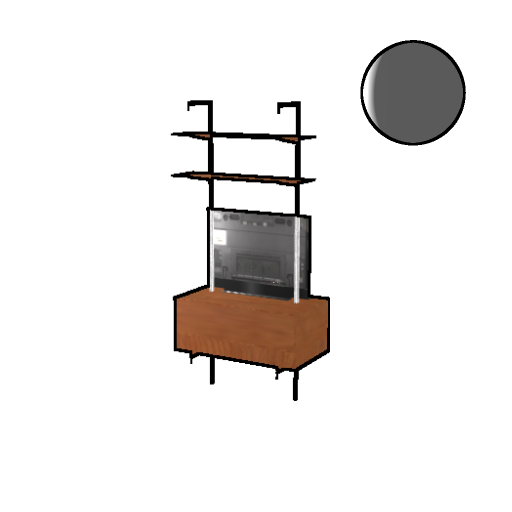}  
& \includegraphics[width=0.12\linewidth,trim={25mm 32mm 15mm 13mm},clip]{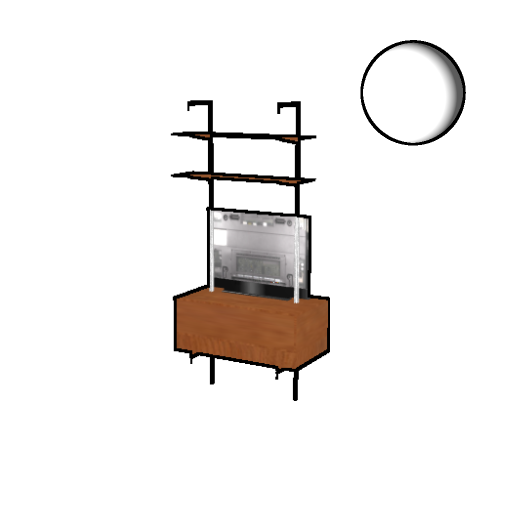}
& \includegraphics[width=0.12\linewidth,trim={25mm 32mm 15mm 13mm},clip]{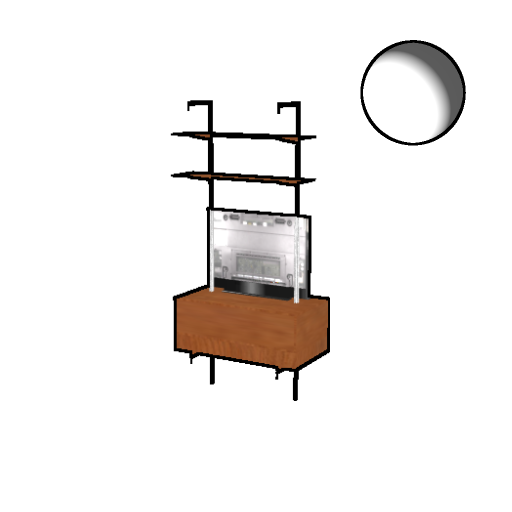}
\\
\includegraphics[width=0.12\linewidth,trim={25mm 32mm 15mm 13mm},clip]{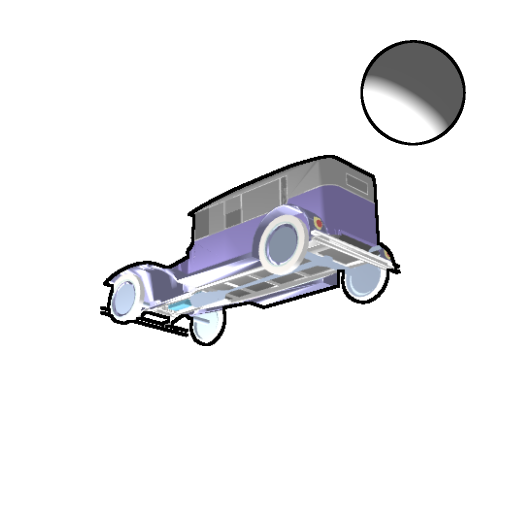}  
& \includegraphics[width=0.12\linewidth,trim={25mm 32mm 15mm 13mm},clip]{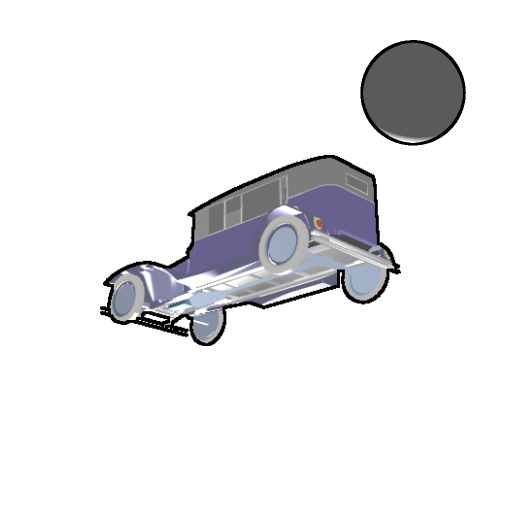}  
& \includegraphics[width=0.12\linewidth,trim={25mm 32mm 15mm 13mm},clip]{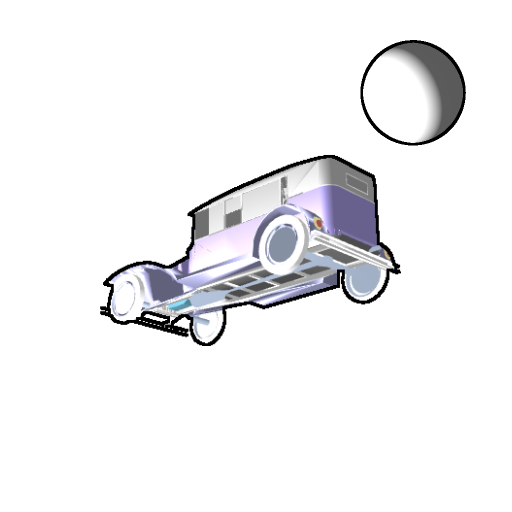}
& \includegraphics[width=0.12\linewidth,trim={25mm 32mm 15mm 13mm},clip]{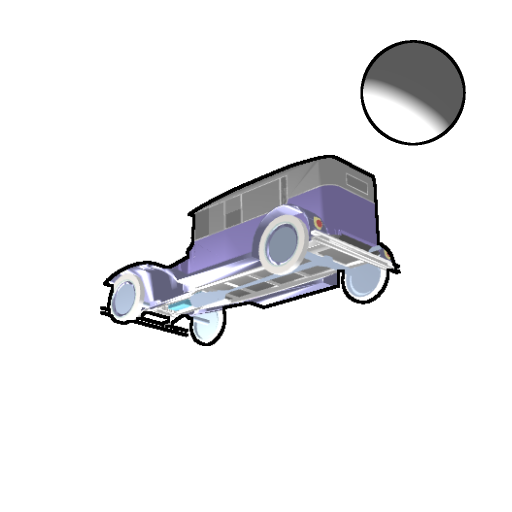}
&\includegraphics[width=0.12\linewidth,trim={25mm 32mm 15mm 13mm},clip]{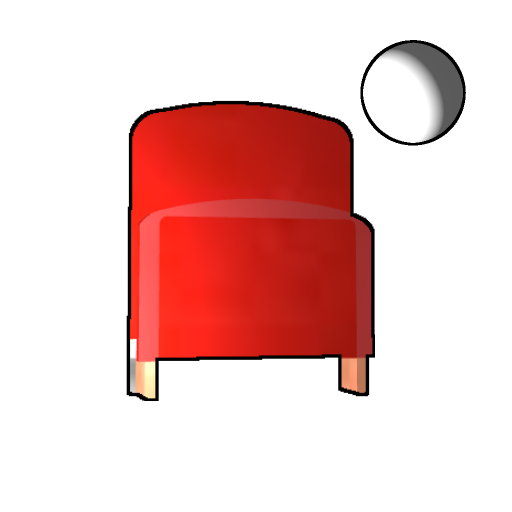}  
& \includegraphics[width=0.12\linewidth,trim={25mm 32mm 15mm 13mm},clip]{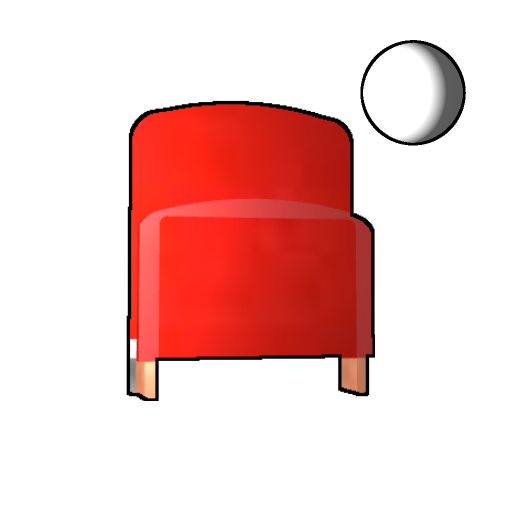}  
& \includegraphics[width=0.12\linewidth,trim={25mm 32mm 15mm 13mm},clip]{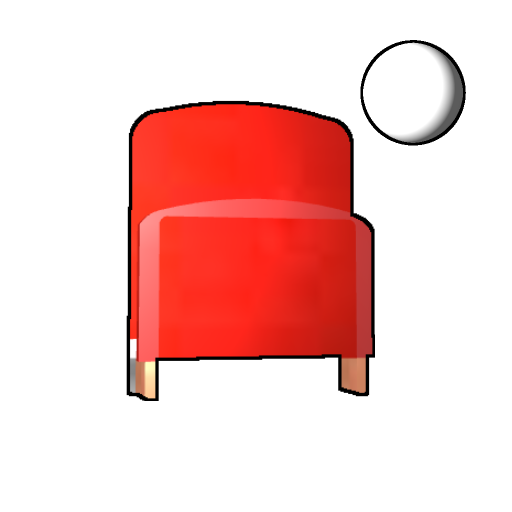}
& \includegraphics[width=0.12\linewidth,trim={25mm 32mm 15mm 13mm},clip]{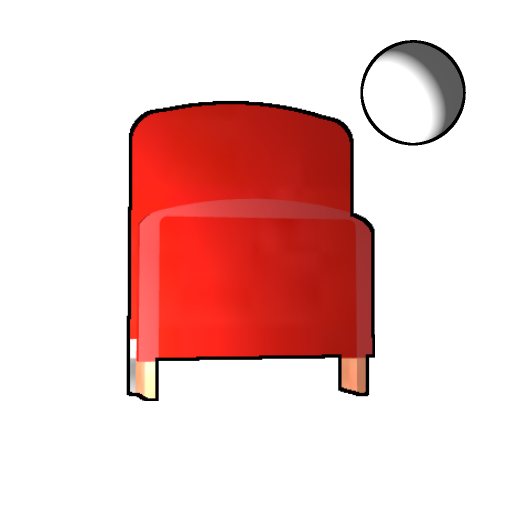}
\\
\includegraphics[width=0.12\linewidth,trim={25mm 32mm 15mm 13mm},clip]{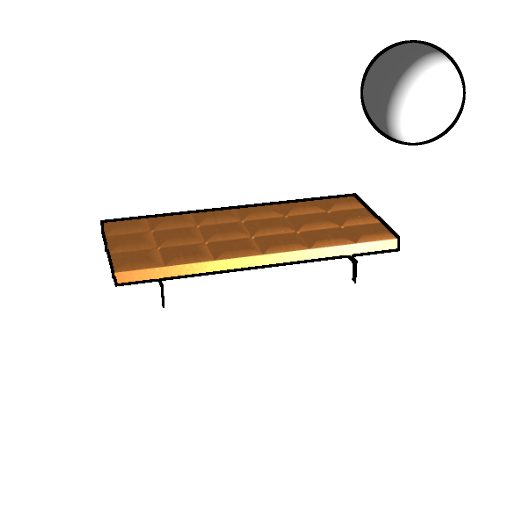}  
& \includegraphics[width=0.12\linewidth,trim={25mm 32mm 15mm 13mm},clip]{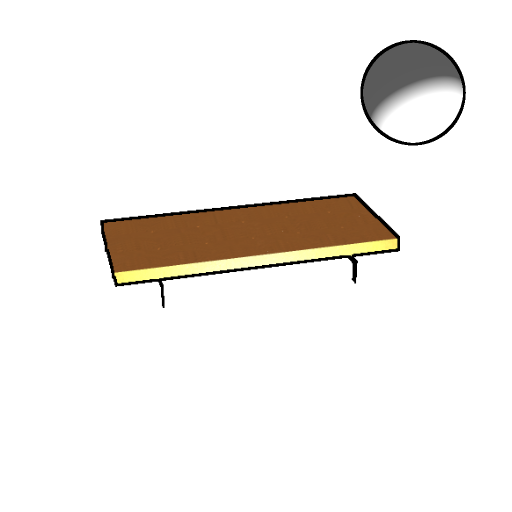}  
& \includegraphics[width=0.12\linewidth,trim={25mm 32mm 15mm 13mm},clip]{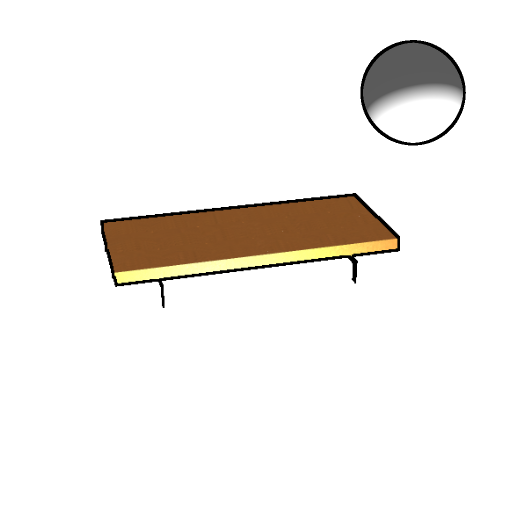}
& \includegraphics[width=0.12\linewidth,trim={25mm 32mm 15mm 13mm},clip]{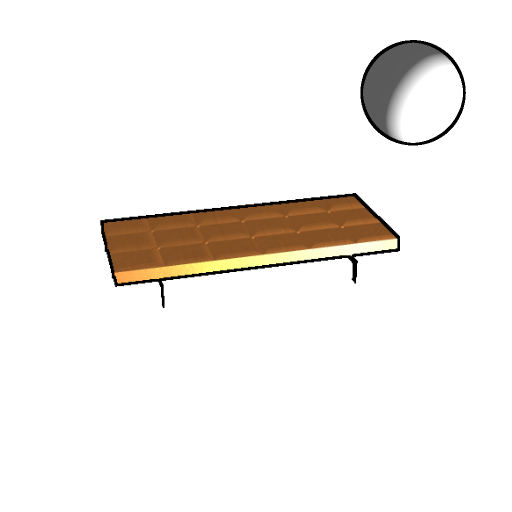}
&\includegraphics[width=0.12\linewidth,trim={25mm 32mm 15mm 13mm},clip]{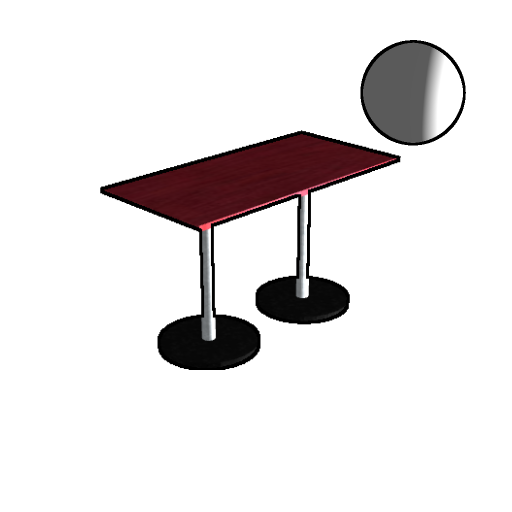}  
& \includegraphics[width=0.12\linewidth,trim={25mm 32mm 15mm 13mm},clip]{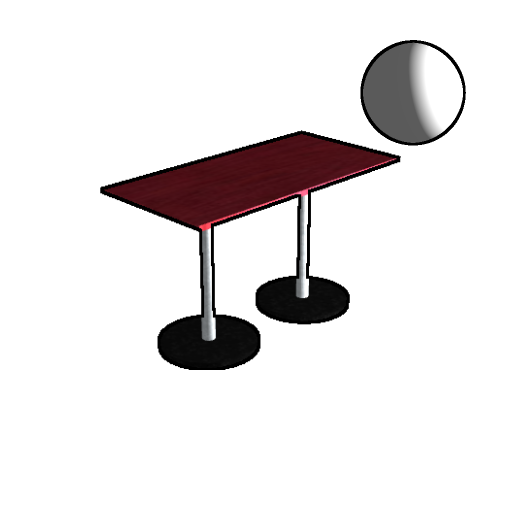}  
& \includegraphics[width=0.12\linewidth,trim={25mm 32mm 15mm 13mm},clip]{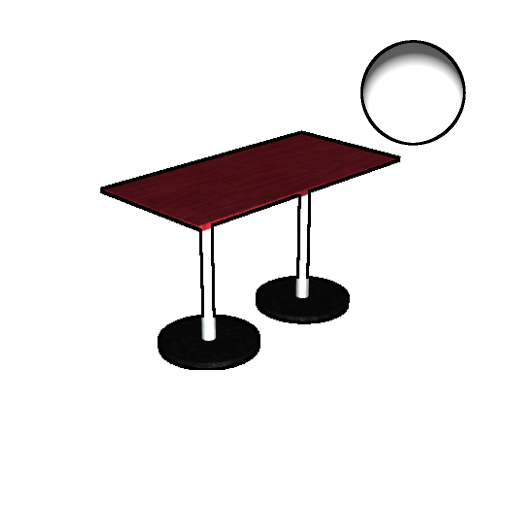}
& \includegraphics[width=0.12\linewidth,trim={25mm 32mm 15mm 13mm},clip]{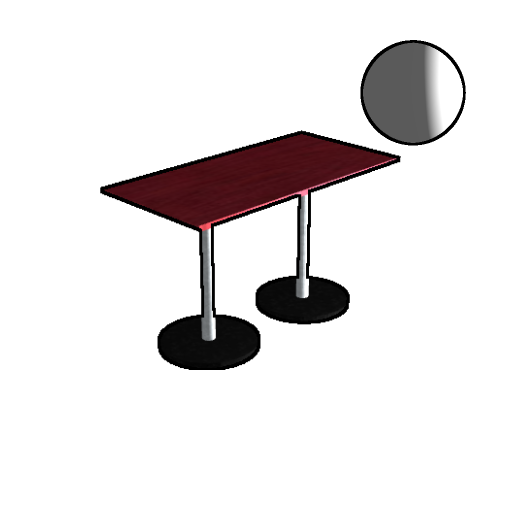}

\end{tabular}
}
\arxiv{-0.3cm}
\caption{\textbf{Qualitative comparison on illumination estimation.} 
}
\arxiv{-0.4cm}
\label{fig:qual-light-comp}
\end{figure*}

\section{Application II: Illumination Estimation}
\arxiv{-3mm}
\paragraph{Problem Formulation:} We next evaluate our method on the task of illumination estimation. The goal is to recover the lighting parameter $\bx \in \mathbb{R}^3$ from the observation RGB image $\by \in \mathbb{R}^{H\times W\times3}$. It has critical applications in image relighting and photo-realistic rendering \cite{karsch2011rendering}. As in the 6-DoF pose estimation task, we assume the 3D model is given. 

\paragraph{Data:} We use the same dataset as the 6-DoF pose estimation experiment for the illumination estimation experiment.  Specifically, we consider two types of light source: directional light and point light. {The two light sources are complementary and can result in very different rendering effect.} During training, we randomly sample the light position from the half unit sphere on the camera side \cite{janner2017self,ma2018single}. If the light is directional, we point the light towards the origin. All the objects are set to have Lambertian surfaces. We ignore the scenario where the light source lies on the other side of the object, as it has no effect on the rendered image. For evaluation, we follow the same criteria. We perform rendering in pyrender and the image size is set to $256 \times 256$. Empirically we found this size provides the best balance between performance and the computational speed.

\paragraph{Metrics:} Following \cite{janner2017self}, we use the standard mean-squared error (MSE) between the ground truth light and estimated light pose to measure the difference. We also compute the outlier rate as described in Sec. \ref{sec:pose-estimation}. \todo{mention what is the threshold for direction light and threshold for positional light}

\paragraph{Network architecture:} 
We employ an encoder-decoder architecture with skip connections as our deep feedback network. Since the 3D geometry of the object plays an important role during rendering, we adopt depth prediction as an auxiliary task. This allows the model to implicitly capture such notion and reason about its relationship with illumination. During training, our deep feedback network estimates both the depth of the object as well as the illumination parameters. 
We use MSE as the objective for both tasks. During inference, we simply discard the depth decoder and output only the illumination part. We unroll our network for 7 steps according to the validation performance.%

\paragraph{Baselines:} 
We exploit NMR \cite{kato2018neural} to minimize the energy $E_\mathrm{data} + E_\mathrm{prior}$.%
The data term is the $\ell_2$ distance between the observation image and the rendered image, while the prior term constrains the light source to lie on the sphere. 
We adopt the same stopping criteria as in Sec. \ref{sec:pose-estimation}.
The size of the rendered image is set to $256 \times 256$ based on the performance on the validation set. For deep regression method, we exploit the state-of-the-art model from Janner \etal \cite{janner2017self}.

\paragraph{Results:} As shown in Tab. \ref{tab:light-quant}, our deep feedback network outperforms the baselines on both setup. The improvement is significant especially in the directional light case. We conjecture this is because the intensity of directional light does not decay w.r.t. the travel distance, and  the signals from the image are weaker. Learning based approaches thus have to rely on feedback signals to refine the light direction. 
The performance of the optimization method is limited by the hand-crafted energy as well as the capability of renderer. NMR is sub-optimal as it approximates the gradient with a manually designed function and does not handle self-occlusion. In contrast, our method allows us to exploit complex rendering machines as the forward model as we do not require it to be differentiable.
We note that we only report the optimization results on directional light since NMR does not support point light source. Fig.~\ref{fig:qual-light-comp} depicts the qualitative comparison against the baselines. It is clear that our deep feedback mechanism is able to recover accurate lighting information based on subtle difference between the forward results and the observations. 

\begin{table}[tb]
\centering
\scalebox{0.95}{
\begin{tabular}{lcccccccccc}
\specialrule{.2em}{.1em}{.1em}
& \multicolumn{2}{c}{Optimization} &\multicolumn{2}{c}{Position Error (cm)} &\multicolumn{2}{c}{Rotation Error ($^{\circ}$)} \\
Methods & Step & Time & Mean & Median & Mean & Median\\
\hline
L-BFGS \cite{grochow2004style} &73  &27.9 s & 0.38 & 0.01 & 7.19 &4.68\\
Adam \cite{kingma2014adam} &196 &38.8 s & 0.04 &0.04 &7.96 & 7.92 \\
Deep6D \cite{zhou2019continuity} &1 & 0.012 s & 1.9 & 1.6 & - & -\\
Ours &4 & 0.12 s& 0.64& 0.36 & 0.88 & 0.03\\
\specialrule{.1em}{.05em}{.05em}
\end{tabular}
}
\caption{\textbf{Quantitative results on CMU MoCap.} }
\label{tab:cmu-quant}
\arxiv{-4mm}
\end{table}

\begin{figure*}[tb]
\centering
\setlength{\tabcolsep}{1pt}
\arxivscale{0.98}{
\begin{tabular}{cccccc}
GT & Step 1 & Step 3 &GT & Step 1 & Step 3\\
\includegraphics[width=0.16\linewidth,trim={57mm 40mm 55mm 35mm},clip]{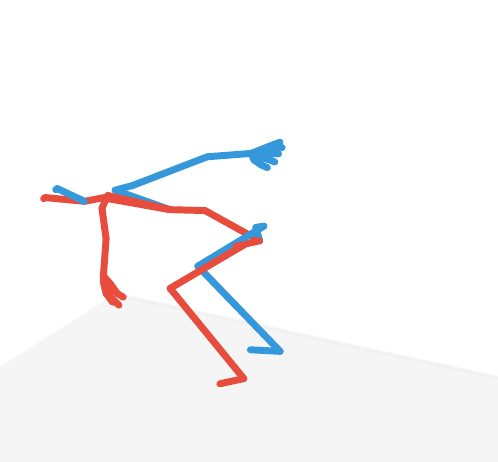}  
&\includegraphics[width=0.16\linewidth,trim={57mm 40mm 55mm 35mm},clip]{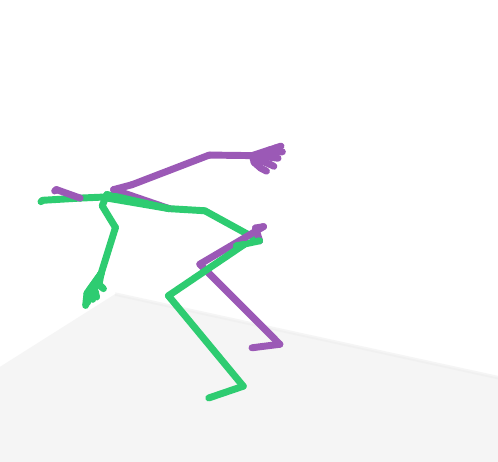}  
&\includegraphics[width=0.16\linewidth,trim={57mm 40mm 55mm 35mm},clip]{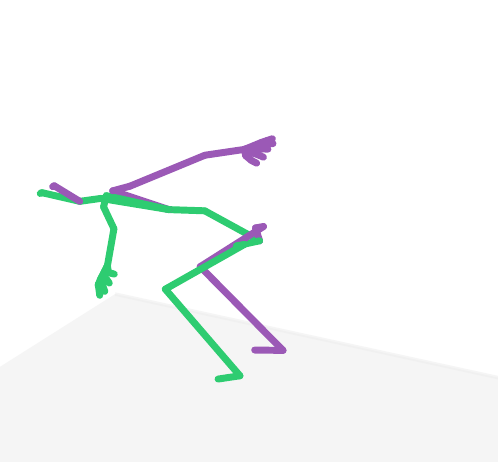} 
&\includegraphics[width=0.16\linewidth,trim={57mm 40mm 55mm 35mm},clip]{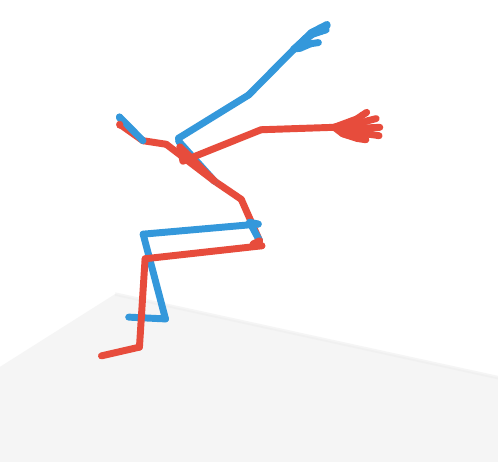}  
&\includegraphics[width=0.16\linewidth,trim={57mm 40mm 55mm 35mm},clip]{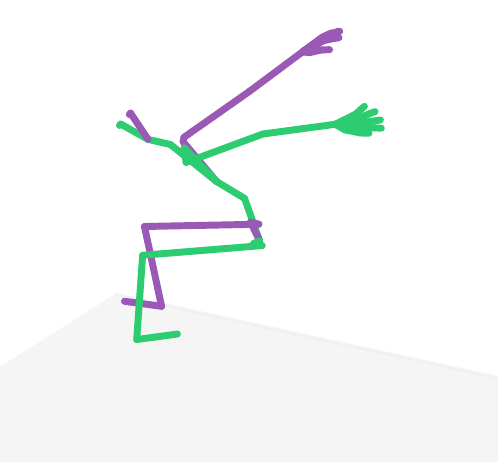}  
&\includegraphics[width=0.16\linewidth,trim={57mm 40mm 55mm 35mm},clip]{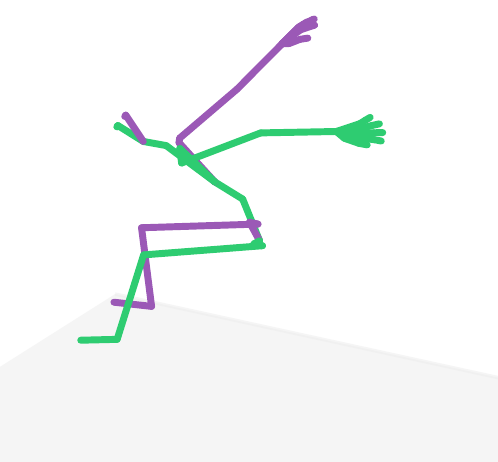} \\
\includegraphics[width=0.16\linewidth,trim={57mm 40mm 55mm 35mm},clip]{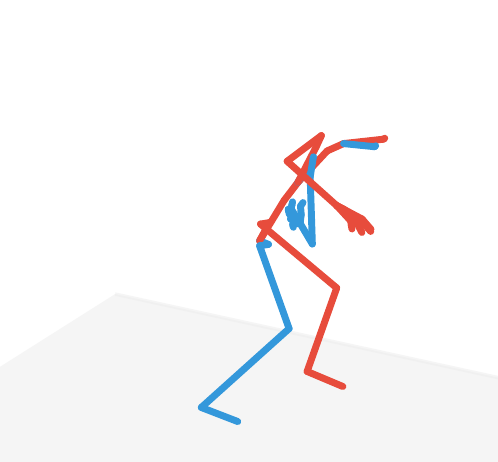}  
&\includegraphics[width=0.16\linewidth,trim={57mm 40mm 55mm 35mm},clip]{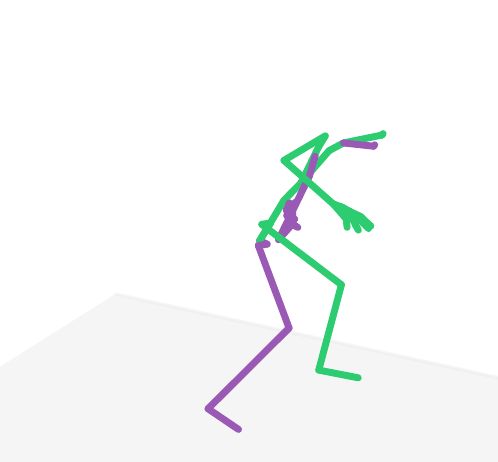}  
&\includegraphics[width=0.16\linewidth,trim={57mm 40mm 55mm 35mm},clip]{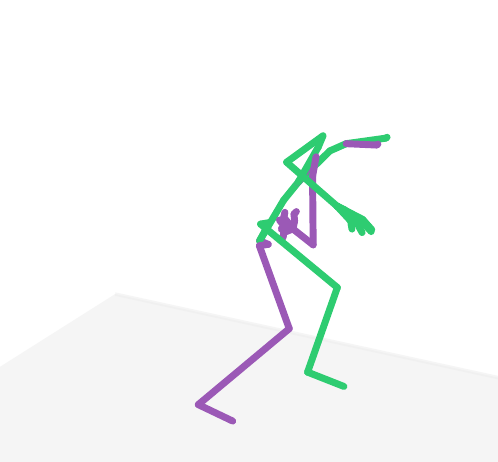} 
&\includegraphics[width=0.16\linewidth,trim={57mm 40mm 55mm 35mm},clip]{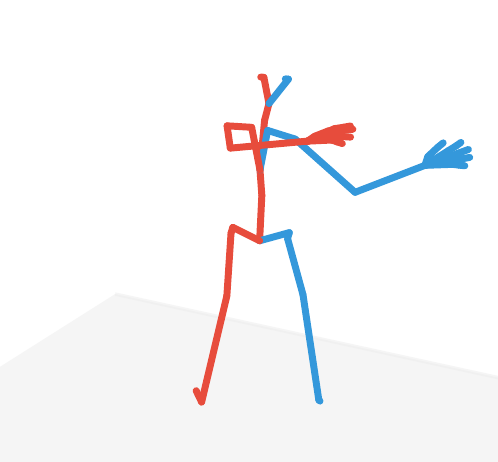}  
&\includegraphics[width=0.16\linewidth,trim={57mm 40mm 55mm 35mm},clip]{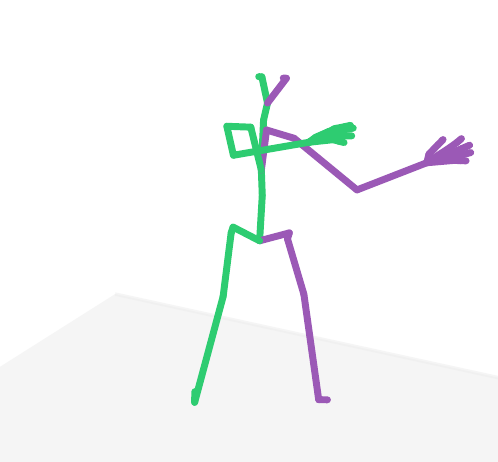}  
&\includegraphics[width=0.16\linewidth,trim={57mm 40mm 55mm 35mm},clip]{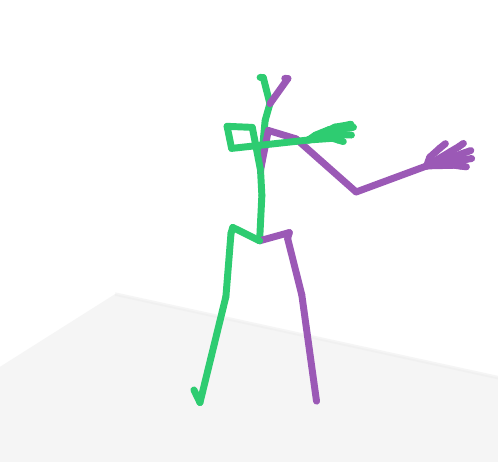} \\
\includegraphics[width=0.16\linewidth,trim={57mm 40mm 55mm 35mm},clip]{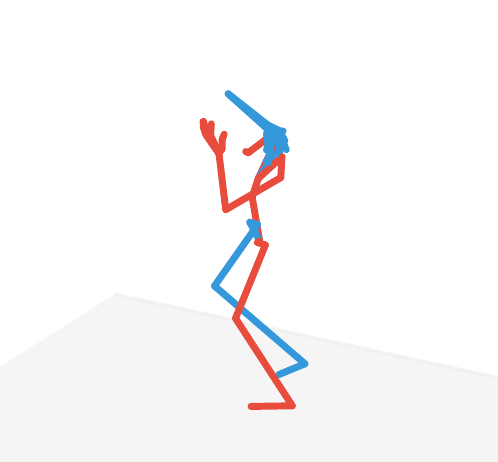}  
&\includegraphics[width=0.16\linewidth,trim={57mm 40mm 55mm 35mm},clip]{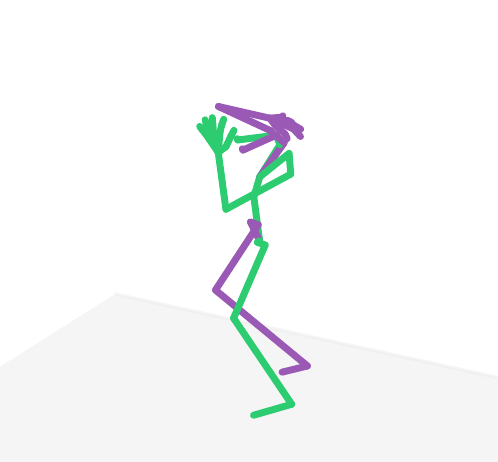}  
&\includegraphics[width=0.16\linewidth,trim={57mm 40mm 55mm 35mm},clip]{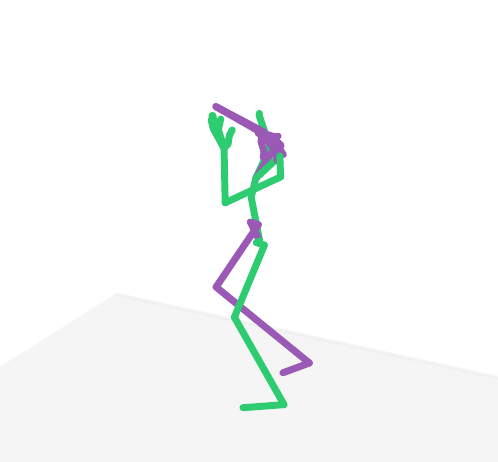} 
&\includegraphics[width=0.16\linewidth,trim={57mm 40mm 55mm 35mm},clip]{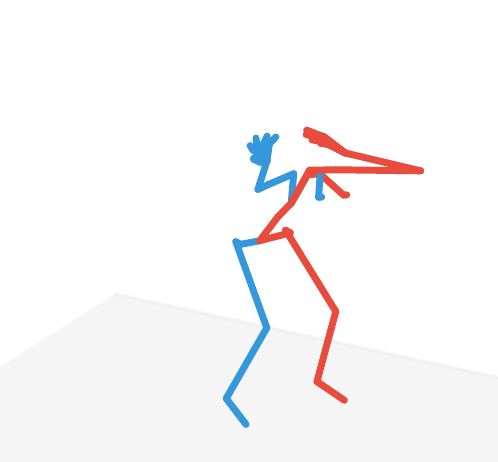}  
&\includegraphics[width=0.16\linewidth,trim={57mm 40mm 55mm 35mm},clip]{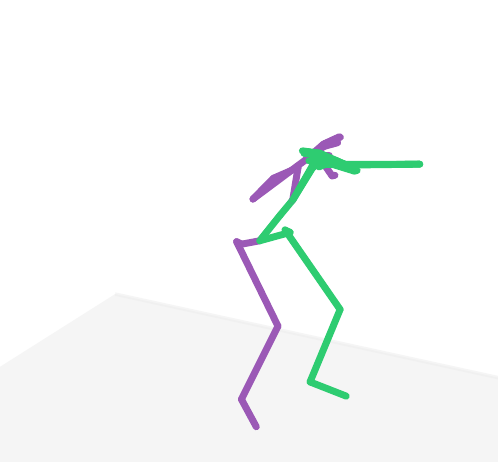}  
&\includegraphics[width=0.16\linewidth,trim={57mm 40mm 55mm 35mm},clip]{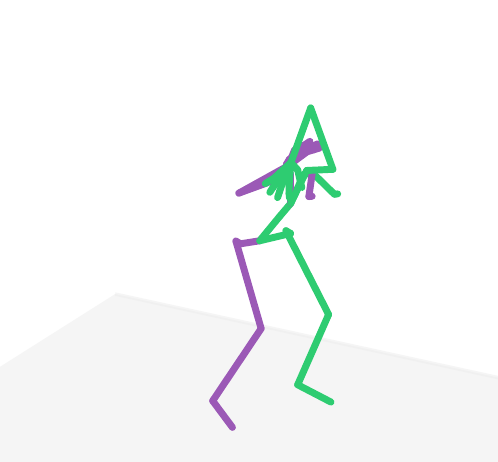} \\
\end{tabular}
}
\arxiv{-0.3cm}
\caption{\textbf{Qualitative results on CMU MoCap:} Our approach is able to accurately predict the joint rotations within a few steps. It can also correct wrong estimations through the feedback from the forward model (see the feet/toes in the right column). Bottom right shows an example where our model fails.}
\arxiv{-0.3cm}
\label{fig:qual-ik}
\end{figure*}

\section{Application III: Inverse Kinematics}

\paragraph{Problem formulation} 
Finally we exploit how our proposed method to tackle the inverse kinematics problem.
Given the 3D location of the joints of a reference pose $\by_{1:N}^{\text{ref}}$ and the desired joint rotations $\bx_{1:N} \in$ SO(3), the forward kinematics function $f$ rotates the joints and computes their 3D positions by recursively applying the follow update rule from parents to children: $
\by_n = \by_{\text{parent}(n)}+ \bx_{n} (\by_{n}^{\text{ref}} - \by_{\text{parent}(n)}^{\text{ref}})$.
The goal of inverse kinematics is to recover the SO(3) rotations $\bx_{1:N}$ that ensure the specific joints are placed at the desired 3D locations $\by_{1:N}$.
Inverse kinematics has a wide range of applications, such as robot arm manipulation, legged robot motion planning and computer re-animation. 
The problem is inherently ill-posed as different rotations can result in the same observation through the forward kinematics function  $f$, \ie, $\by = f(\bx_{1:N}) = f(\bx_{1:N}^{\prime})$. However, not all angles are feasible or natural due to the dynamic constraints.
Therefore, in order to accurately recover the rotations, one has to either come up with a powerful prior or learn it from data.
In this paper, we focus on inverse kinematics over human body skeletons. 

\paragraph{Data:} We validate our model on the CMU Motion Capture Dataset (CMU MoCap) as it contains complex human motions and a diverse range of joint rotations. Following Yi \etal \cite{zhou2019continuity}, we select 865 motion clips from 37 motion categories and hold out 37 clips for testing. Each skeleton in the dataset has 57 joints. We fix the position of the hip to remove the effect of global motion.

\paragraph{Metrics:} We evaluate the performance of our model with joint position error \cite{zhou2019continuity} and joint angular error \cite{martinez2017human,pavllo:quaternet:2018}. The two metrics are complementary since a small rotation error may result in a large position error due to the recursive nature of the forward kinematics model, and small position error cannot guarantee correct joint rotation due to ambiguities.

\paragraph{Network architecture:}
Our deep feedback network is a multilayer perception akin to \cite{zhou2019continuity}. Following \cite{villegas2018neural,pavllo:quaternet:2018}, the network takes as input the estimated joint position, reference joint position, as well as the difference between the two, and outputs a rotation for each joint. We unroll our model three steps. We train the network with L2 loss on both position error and rotation error. 

\paragraph{Baselines:} We compare our model against two optimization-based approaches and one deep regression method. 
For optimization methods, we employ joint position error as our data term, \ie $E_{\text{data}}(f(\bx), \by) = \lVert f(\bx) - \by \rVert_2^2$, and derive a prior energy term from data to alleviate the ambiguities of joint rotations. In particular, we fit a gaussian distribution over the Euler angles of each joint from training data and employ it as a regularization term during inference. 
We set the weight of the prior term to $0.001$ and optimize both energies jointly. 
We exploit two different types of optimizers: a first-order method (\ie, Adam \cite{kingma2014adam}) and a quasi-Newton method (\ie, L-BFGS \cite{grochow2004style}). 
For deep regression method, we compare with the current state of the art (Deep6D \cite{zhou2019continuity}). 
\paragraph{Results:} As shown in Tab. \ref{tab:cmu-quant}, our deep feedback network outperforms the baselines on the rotation metric and achieve comparable performance on the position error.
By unrolling more steps and gathering feedback signals from the forward model, we are able to reduce incorrect estimation and improve the performance (see the Fig. \ref{fig:qual-ik}). We refer the readers to the supp. material for detailed analysis.
On average, a single step of L-BFGS, Adam, and our approach takes 383 ms, 198 ms, 30 ms respectively. L-BFGS takes longer to compute as it needs to conduct gradient evaluation multiple times to approximate the Hessian. Adam is faster in terms of computation, yet it takes far more steps to converge.  Our approach, in comparison, is significantly faster and better. 

\todo{explain why Deep6D doesn't have rotation error.}

%
%
%
%
%
%

%
%

%

%
%
%
%

%
%

%
%

%
%
%
%
%
%
%
%
%
%
%
%

%
%
%
%
%
%
%
%
%
%
%
%
%
%
%
%
%
%

%

%
%
%
%
%
%
%
%
%
%
%
%
%
%
%
%
%
%
%

%
%
%
%
%
%
%
%
%
%
%
%
%
%
%
%
%
%
%
%
%
%
%
%
%
%
%
%
%

%
%
%
%
%
%
%
%
%
%
%
%
%
%
%
%
%
%
%
%
%
%
%
%
%
%
%
%

%
%
%
%
%
%
%
%
%
%
%
%

%
%
%
%
%
%
%
%
%
%
%
%
%
%
%
%
%
%
%
%
%
%
%
%
%
%
%
%
%

%
%
%
%
%
%
%
%
%
%
%
%
%
%
%
%
%
%
%
%
%
%
%
%
%

%
%
%
%
%
%
%
%
%
%
%
%
%
%
%
%
%
%

%
%
%
%
%
%
%
%
%
%
%
%
%
%
%
%

%
%
%
%
%
%
%
%
%
%
%
%
%
%
%
%

%
%
%
%
%
%
%
%
%
%
%
%
%
%
%

%
%
%
%
%
%
%
%
%
%
%
%

%
%
%
%
%
%
%
%
%
%
%
%

%
%
%
%
%
%
%
%
%
%
%
%
%
%
%
%
%
%
%
%
%
%
%
%
%
%
%
%
%
%
%
%
%
%
%
%
%
%
%
%
%
%
%
%
%
%
%
%
%
%
%
%
%
%
%
%
%
%
%
%
%
%
%
%
%
%
%
%
%
%
%
%
%
%
%
%
%
%
%
%
%
%
%
%
%
%
%
%
%
%
%

%
%
%
%
%
%
%
%
%
%
%
%
%
%
%
%
\arxiv{-3mm}
\section{Conclusions}
\arxiv{-3mm}
In this paper, we propose a deep feedback inverse problem solver. Our method combines the strength of both learning-based approaches and optimization-based methods. Specifically, it learns to conduct an iterative update over the current solution based on the feedback signals provided from the forward process of the problem. Unlike prior work, it does not have any restrictions on the forward process. Further, it learns to conduct an update without explicitly define an objective function. Our results showcase that the proposed method is extremely effective, efficient, and widely applicable.

{\small \paragraph{Acknowledgement} The authors would like to thank Ching-Yao Chuang, Frieda Rong, Mengye Ren for their feedback on the early draft, Jonas Wulff, Tongzhou Wang for their helpful disccusions, Chris Williams, Andrea Tagliasacchi for bringing up relevant literature, Yen-Chen Lin for pointing out the misprint in the figures, and the anonymous reviewers for their constructive suggestions.}

{%
\bibliographystyle{splncs04}
\bibliography{egbib}
}

\end{document}